\documentclass[lettersize,journal]{IEEEtran}   

\IEEEoverridecommandlockouts                              

\title{\LARGE \bf
Sphere-VIO: Fast and Robust Visual-Inertial Odometry via Unified Spherical Representation for Heterogeneous Multi-Camera Systems}

\author{Yueteng Yang$^{1}$, Yusen Xie$^{1}$, Hao Wei$^{2}$, Qianhao Wang$^{3}$, Boyu Zhou$^{4}$, Fei Gao$^{3}$, Jun Ma$^{1}$, and Jinni Zhou$^{1}$
\thanks{$^{1}$Yueteng Yang, Yusen Xie, Jun Ma, and Jinni Zhou are with The Hong
Kong University of Science and Technology (Guangzhou), Guangzhou 511453, China (e-mail: yyang873@connect.hkust-gz.edu.cn;  yxie827@connect.hkust-gz.edu.cn; jun.ma@ust.hk; eejinni@connect.hkust-gz.edu.cn)}%
\thanks{$^{2}$Hao Wei is with Differential Robotics, Hangzhou 311100, China (e-mail: weihao@diffrobot.com)}%
\thanks{$^{3}$Qianhao Wang and Fei Gao are with 
Zhejiang University, Hangzhou 310027, China (e-mail: qhwangaa@zju.edu.cn; fgaoaa@zju.edu.cn)%
}
\thanks{$^{4}$Boyu Zhou is with Southern University of Science and Technology, Shenzhen 518055, China         (e-mail: zhouby@sustech.edu.cn)%
}
}

\usepackage{amsmath}
\usepackage{mathtools}
\usepackage{comment}
\usepackage{amssymb}
\usepackage{threeparttable} 
\usepackage{multirow}       
\usepackage{booktabs}       
\usepackage{siunitx}  
\usepackage{graphicx}    
\usepackage{subcaption}
\usepackage{tikz}
\usepackage{soul}
\usepackage{booktabs}
\usepackage[table,xcdraw]{xcolor}
\newcommand{\colorfirst}{\cellcolor[HTML]{c0e2ca}}
\newcommand{\colorsecond}{\cellcolor[HTML]{fff5b3}}
\newcommand{\colorthird}{\cellcolor[HTML]{ffd9b3}}

\newcommand{\boxfirst}{\colorbox[HTML]{c0e2ca}}
\newcommand{\boxsecond}{\colorbox[HTML]{fff5b3}}
\newcommand{\boxthird}{\colorbox[HTML]{ffd9b3}}
\begin{document}

\maketitle
\thispagestyle{empty}
\pagestyle{empty}

\begin{abstract}

Multi-camera visual-inertial odometry (VIO) overcomes the inherent limitations of pure visual systems by expanding the field of view. However, existing algorithms are typically tailored for fixed camera setups and lack unified compatibility with heterogeneous multi-camera systems. Meanwhile, due to the absence of a unified cross-camera representation and association mechanism, current methods struggle to achieve a balance among robust cross-camera feature tracking, stable depth estimation, and reliable real-time performance. To address these issues, we present Sphere-VIO, a lightweight filter-based VIO framework with unified spherical representation for heterogeneous multi-camera systems. Specifically, we first propose a Unified Spherical Panorama Model (USPM) that supports all standard camera models and enables bidirectional fast mapping between multi-camera images and a shared spherical space without sequential stitching, simplifying cross-camera feature management and improving triangulation efficiency. Second, we design a parallel-accelerated depth-guided semi-direct tracking pipeline, namely Hierarchical Omnidirectional Feature Alignment (HOFA), with global spherical constraints for robust cross-camera matching, and fuse multi-camera depth observations into a standard depth filter for stable initialization. Finally, we develop a multi-camera-adapted ESKF backend that employs spherical bearing residuals and Schur complement marginalization to minimize computational overhead, enabling accurate real-time state estimation on resource-constrained devices. Extensive experiments on public benchmarks and a custom omnidirectional dataset show that Sphere-VIO achieves superior trade-offs between accuracy, robustness, efficiency, and cross-camera generality.

\end{abstract}

\section{Introduction}
\label{sec:intro}
Pure visual methods suffer from severe performance degradation under illumination changes, rapid motion, and texture-less regions, driving the adoption of multi-camera systems for enhanced robustness. With wider Field of View (FOV), stronger observational stability, and complementary perception, multi-camera systems have become mainstream hardware for robust localization in challenging real scenes. However, existing methods have key limitations: a series of methods \cite{yang2017multi,he2022towards} 
track features per-camera without global cross-camera association, 
while others 
\cite{seok2019rovo,pan_slam,mcov_slam}
are tailored for specific setups such as \ang[round-mode=none]{360} fisheye rigs and lack generality for arbitrary heterogeneous multi-camera modules.
Visual-inertial odometry (VIO) fuses visual and IMU data for better robustness, but still suffers from the above multi-camera processing defects: brute-force overlapped view processing causes excessive redundancy \cite{mavis}, while divergent orientations and severe distortion exacerbate cross-camera alignment difficulties \cite{rovins,d2slam}.
Single-lens panoramic cameras \cite{panoair,360_vio,lfvislam} provide full FOV but lack baseline constraints for reliable scale estimation, forcing reliance on noisy IMU scale cues and leading to unstable scale and long-term drift. Recent works \cite{pan_slam,multi_lvi_sam} adopt panoramic representations, a special full-field variant of general spherical representations, but remain confined to specific rigid setups, ignore geometric consistency for heterogeneous modules, and depend on expensive stitching or optimization backends, resulting in poor real-time performance. Therefore, balancing generality, lightweight efficiency, tracking robustness, and estimation accuracy remains a critical challenge for arbitrary multi-camera and panoramic VIO systems.
To address these challenges, this paper proposes Sphere-VIO, a filter-based visual-inertial odometry framework for arbitrary heterogeneous multi-camera modules, powered by a unified spherical representation. The framework follows a three-step pipeline. First, we construct the Unified Spherical Panorama Model (USPM) that enables bidirectional fast mapping between multi-camera images and a shared spherical space, supporting arbitrary intrinsic types without sequential stitching or fusion. Second, based on this unified representation, we implement a Hierarchical Omnidirectional Feature Alignment (HOFA) procedure that combines epipolar constraints with adaptive depth range filtering under spherical constraints, followed by multi-camera joint depth observation and a standard depth filter for stable depth estimates. Third, we design a multi-camera-adapted Error State Kalman Filter (ESKF) backend that uses spherical bearing vector residuals as observations and marginalizes landmarks via Schur complement, ensuring lightweight CPU computation. The experiment on the challenging HILTI 2022 dataset shows that our method achieves an average ATE RMSE of 0.1821 m with an average per-frame processing time as only 0.0122 s, outperforming all compared stereo visual SLAM methods in localization accuracy and efficiency. The core contributions are as follows:
\begin{itemize}
    \item We propose the Unified Spherical Panorama Model that supports arbitrary camera models, eliminates sequential stitching, and provides consistent cross-camera feature representation with efficient triangulation.
    \item We present the Hierarchical Omnidirectional Feature Alignment that integrates epipolar constraints with adaptive depth filtering to achieve robust cross-camera matching and joint depth estimation, eliminating redundancy and mismatches in overlap.
    \item We develop a Multi-Camera-Adapted ESKF Backend that uses spherical bearing residuals and Schur-complement marginalization for low-complexity, real-time state estimation using only CPU resources.
    \item We experimentally showcase that our VIO framework supports various camera setups and achieves low computation overhead plus high robustness in multi-camera scenarios.
\end{itemize}
\section{RELATED WORK}
\subsection{Forward Stereo Visual SLAM}
Forward stereo is the most classical and widely used setup for visual SLAM and VIO. With strict epipolar constraints, they enable direct metric scale recovery and stable state estimation, solving the inherent scale ambiguity of monocular systems. Representative methods consist of feature-based frameworks such as ORB-SLAM3 \cite {orb_slam3} and VINS-Fusion \cite {qin2018vins}, as well as semi-direct and direct schemes including SVO 2.0 \cite {svo_2_0}, SchurVINS \cite {schurvins} and DSO \cite {dso}. These approaches balance efficiency and accuracy for real-time applications. However, they suffer from three inherent, irreparable limitations: tracking failure under aggressive motion caused by extremely narrow FoV, inflexibility to non-coaxial, heterogeneous or arbitrary multi-camera modules beyond parallel rigid setups, and degraded performance in texture-sparse and occluded regions lacking multi-view complementarity. To address these issues, researchers have increasingly adopted non-coaxial multi-camera systems for wider perceptual range and better robustness.
\subsection{Non-Coaxial Multi-Camera Visual SLAM}
Non-coaxial multi-camera systems offer wider FOV and better robustness than forward-facing stereo setups. Recent generic multi-camera SLAM works \cite{kaveti2023design,cui2023mcsfm,yu2025robust} support arbitrary camera arrangements, but as pure visual systems without IMU fusion, they lack robustness and real-time performance on embedded platforms. CuVSLAM \cite{korovko2025cuvslam} achieves real-time multi-camera localization via CUDA on GPU edge devices, yet it is locked to NVIDIA GPU and cannot run on CPU-only hardware. By configuring cameras as stereo pairs, it only works with pinhole arrays and excludes heterogeneous fisheye and wide-FOV rigs. A multi-camera VIO method \cite{he2022towards} is limited to non-overlapping monocular setups without stereo baselines, leading to unstable scale estimation, and such configurations are less practical than off-the-shelf panoramic cameras. VILENS-MC \cite{zhang2022balancing} eliminates redundant landmarks via cross-camera feature tracking, but its factor graph backend introduces significant overhead and hinders lightweight real-time execution on CPUs. A multi-stereo system \cite{jaekel2020robust} supports arbitrarily oriented stereo pairs, but cannot handle heterogeneous non-parallel multi-camera systems. MAVIS \cite{mavis}, ROVINS \cite{rovins}, and D2SLAM \cite{d2slam} support partially overlapping omnidirectional fisheye rigs, but process each camera individually with heavy optimization-based backends, resulting in inefficient cross-camera feature management.
\begin{figure}[!t]
    \centering
    \includegraphics[width=\linewidth]{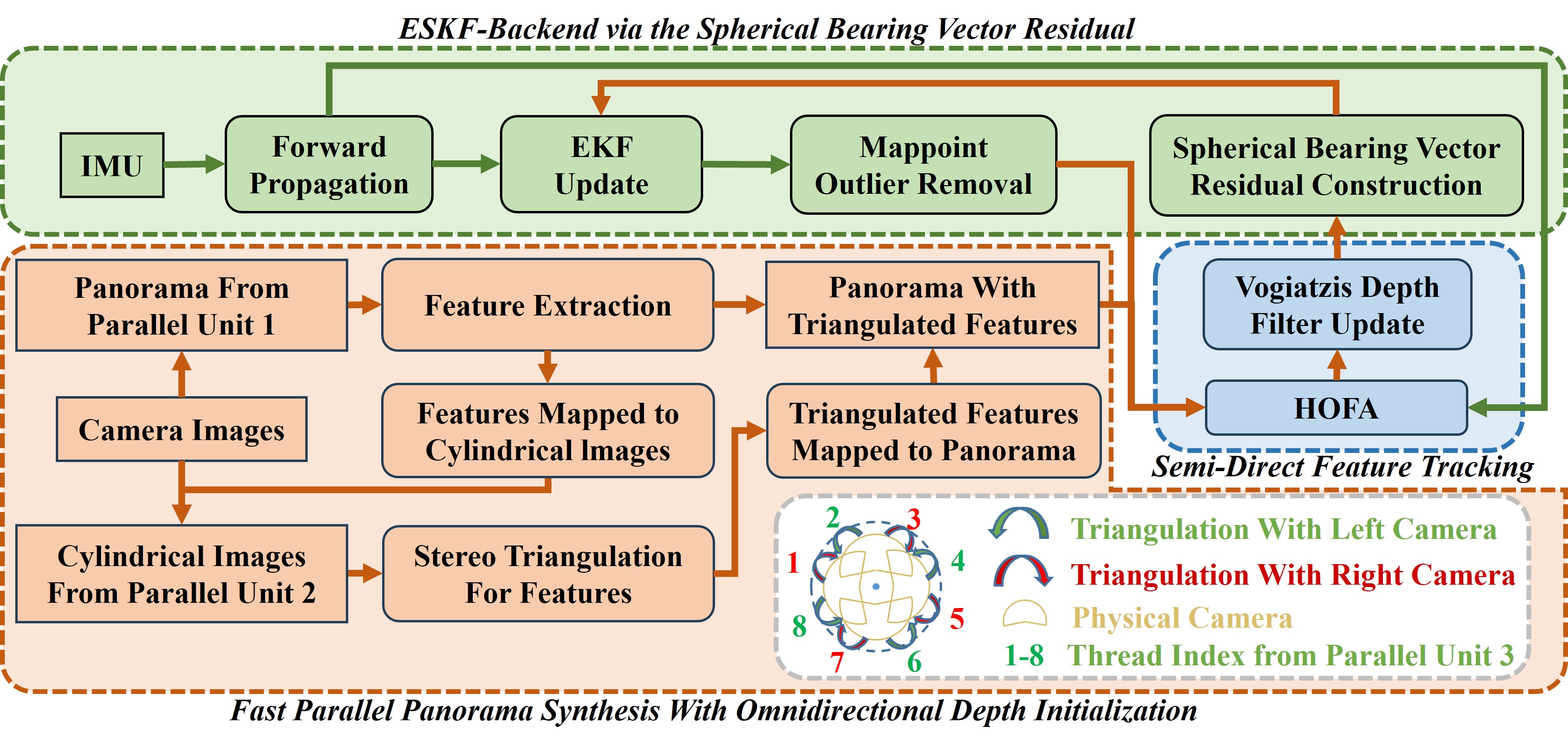}
    \caption{Overview of the Sphere-VIO Framework.
The orange module implements the USPM for multi-camera-to-spherical mapping, the blue module  realizes the HOFA procedure with epipolar constraints and multi-camera depth observation, and the green module uses Schur complement marginalization for lightweight ESKF computation. The inset shows the internal details of stereo triangulation for features.}
    \label{fig:overview_pipeline}
\end{figure}
\section{methodology}
Section \ref{subsec:universal_camera_model} proposes a universal camera model named USPM for fast parallel panorama synthesis and omnidirectional depth initialization. Section \ref{subsec:hofa} describes the HOFA module for robust semi-direct tracking. Section \ref{subsec:spherical_eskf} introduces an efficient ESKF backend via spherical bearing vector residuals. The overall architecture of Sphere-VIO is illustrated in Fig. \ref{fig:overview_pipeline}, with the three core modules marked in orange, blue and green respectively. We elaborate on each module in the following subsections.
\subsection{Universal Camera Model for Fast Parallel Panorama Synthesis and Depth-Guided Panoramic Feature Tracking} \label{subsec:universal_camera_model} 
We propose the USPM for arbitrary multi-camera rigs, unifying all camera images onto a single spherical surface. USPM is flexible: it does not require full \ang[round-mode=none]{360} coverage and even works with narrow-FOV forward stereo rigs.

The USPM projection is formulated as:
\begin{align}
\prescript{P}{}{\boldsymbol{p}} &= \prescript{P}{C}{\boldsymbol{R}}s \prescript{C}{}{\hat{\boldsymbol{p}}} + \prescript{P}{C}{\boldsymbol{t}},
\quad \text{s.t.} \ \left\lVert\prescript{P}{}{\boldsymbol{p}} \right\rVert_2 = r \label{eq:radius_camera_transformation}
\end{align}
where $\prescript{C}{}{\hat{\boldsymbol{p}}}$ is the normalized 3D point in the physical camera frame $C$, rigidly transformed to the virtual panoramic frame $P$ under the spherical constraint $\|\prescript{P}{}{\boldsymbol{p}}\|=r$. To satisfy this constraint, we derive the scale factor $s$ as:
\begin{align}
s 
&=
 \sqrt{   [ (\prescript{P}{C}{\boldsymbol{R}}\prescript{C}{}{\hat{\boldsymbol{p}}})^\top\prescript{P}{C}{\boldsymbol{t}}]^2  - \left\| \prescript{P}{C}{\boldsymbol{t}} \right\|_2^{2} +  r^{2}  } - (\prescript{P}{C}{\boldsymbol{R}}\prescript{C}{}{\hat{\boldsymbol{p}}})^\top\prescript{P}{C}{\boldsymbol{t}}
\label{eq:scale_factor_s}
\end{align}
$\prescript{P}{}{\boldsymbol{p}}$ is then normalized to unit vector $\prescript{P}{}{\hat{\boldsymbol{p}}}$ as:
\begin{align}
\prescript{P}{}{\hat{\boldsymbol{p}}} &= \frac{\prescript{P}{}{\boldsymbol{p}}}{ r } = 
\begin{bmatrix}
\prescript{P}{}{\hat{\boldsymbol{p}}}_{x} \\
\prescript{P}{}{\hat{\boldsymbol{p}}}_{y} \\
\prescript{P}{}{\hat{\boldsymbol{p}}}_{z}
\end{bmatrix} \label{eq:normalized_p_vector} 
\end{align}
and is converted to azimuth $\prescript{P}{}{\theta}$ and elevation $\prescript{P}{}{\phi}$:
\begin{align}
\prescript{P}{}{\theta} &= 
\arctan\left(\frac{\prescript{P}{}{\hat{\boldsymbol{p}}}_{x}}{\prescript{P}{}{\hat{\boldsymbol{p}}}_{z}}\right),\quad
\prescript{P}{}{\phi} 
= \arcsin(\prescript{P}{}{\hat{\boldsymbol{p}}}_{y}) \label{eq:spherical_angles}
\end{align}
Finally $\prescript{P}{}{\boldsymbol{p}}$ is mapped to panoramic pixels:
\begin{align}
\prescript{P}{}{u} &= (\frac{\prescript{P}{}{\theta}}{{\theta}_{h}} + \frac{1}{2}){\prescript{P}{}{W}},\quad
\prescript{P}{}{v} = (\frac{\prescript{P}{}{\phi} }{{\phi}_{v}} + \frac{1}{2} ){\prescript{P}{}{H}} \label{eq:pano_pixels}
\end{align}
where $\theta_h,\phi_v$ are panoramic FOV ranges, and $\prescript{P}{}{W},\prescript{P}{}{H}$ are image dimensions. $\theta_h$ and $\phi_v$ are not restricted to $2\pi$ and $\pi$, enabling adaptation to arbitrary virtual panoramic specifications.

\begin{figure}
    \centering
    \includegraphics[width=1.0\linewidth]{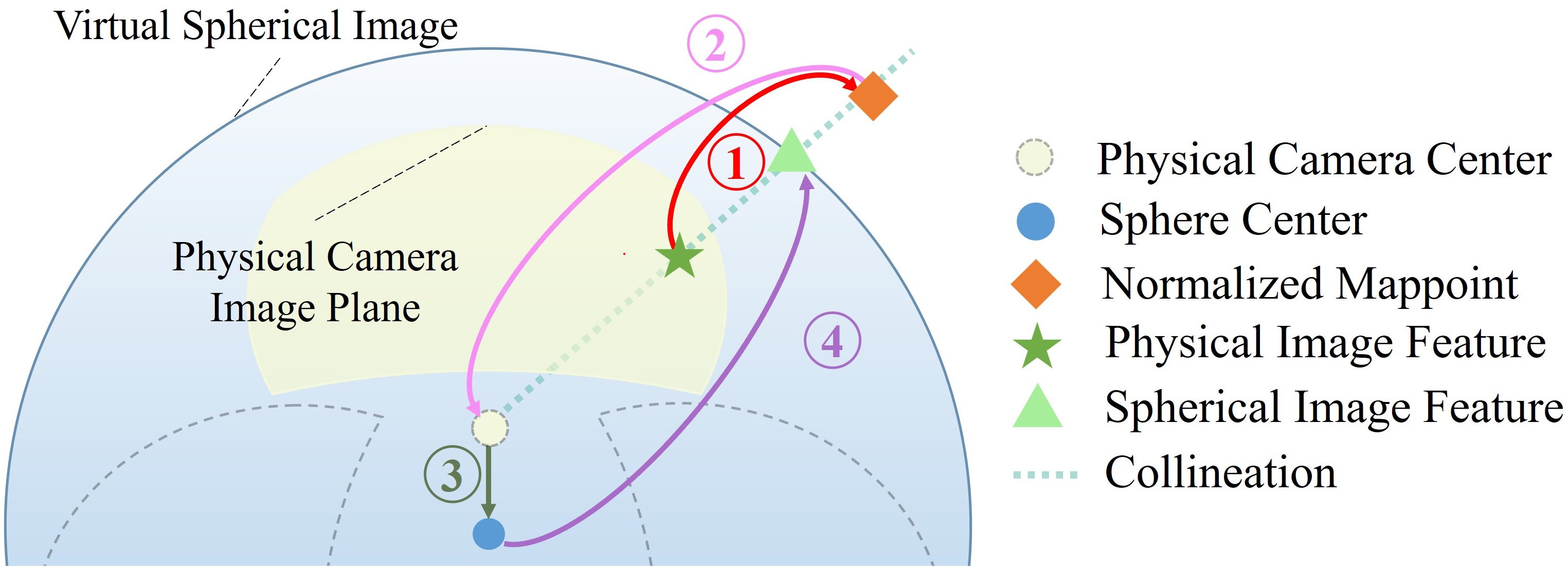}
    \caption{Forward mapping of the proposed USPM. Steps 1-4 illustrate the complete pipeline of projecting physical pixels $^C\boldsymbol{u}$ to panoramic spherical features $^P\boldsymbol{u}$, where Steps 2-4 correspond to the core projection formula of the USPM.}
    \label{fig:USPM_Projection}
\end{figure}

We define this projection into a unified operator $\prescript{U}{}{\boldsymbol{\pi}}(\cdot)$:
\begin{align}
\label{eq:uspm_projection_symbol}
\prescript{P}{}{\mathbf{u}} = \prescript{U}{}{\boldsymbol{\pi}} \left( \prescript{C}{}{\hat{\boldsymbol{p}}} \right)
\end{align}
where $\prescript{P}{}{\mathbf{u}} = [\prescript{P}{}{u}\ \prescript{P}{}{v}]^\top$ is the panoramic pixel vector. This compact notation, illustrated in Fig. \ref{fig:USPM_Projection}, is used throughout subsequent derivations.


Inverse projection is essential for multi-camera panoramic remapping via our USPM. Forward projection $\prescript{U}{}{\boldsymbol{\pi}}(\cdot)$ maps normalized 3D point $\prescript{C}{}{\hat{\boldsymbol{p}}}$ in the physical frame to panoramic pixel $\prescript{P}{}{\boldsymbol{u}}$ in the virtual frame, and its inverse recovers the corresponding 3D direction for cross-camera feature alignment.
Following USPM's forward logic, we derive its geometrically consistent inverse operator:
\begin{align}
\prescript{C}{}{\hat{\boldsymbol{p}}} = {\prescript{U}{}{\boldsymbol{\pi}}}^{-1}\!\left( \prescript{P}{}{\boldsymbol{u}} \right)
\end{align}
with explicit formulation:
\begin{equation}
\begin{aligned}
\prescript{P}{}{\theta} &= \frac{2\prescript{P}{}{u} - \prescript{P}{}{W}}{2\prescript{P}{}{W}}  {{\theta}_{h}}, \quad
\prescript{P}{}{\phi} = \frac{2\prescript{P}{}{v} - \prescript{P}{}{H}}{2\prescript{P}{}{H}}{{\phi}_{v}} \\
\prescript{P}{}{\hat{\boldsymbol{p}}}_{x} &= |\cos(\prescript{P}{}{\phi})|  \cos(\prescript{P}{}{\theta}), \quad
\prescript{P}{}{\hat{\boldsymbol{p}}}_{y} = \sin(\prescript{P}{}{\phi}) \\
\prescript{P}{}{\hat{\boldsymbol{p}}}_{z} &= |\cos(\prescript{P}{}{\phi})|  \sin(\prescript{P}{}{\theta}) \\
\prescript{C}{}{\hat{\boldsymbol{p}}} &= 
\frac{\prescript{C}{P}{\boldsymbol{R}}r \prescript{P}{}{\hat{\boldsymbol{p}}} + \prescript{C}{P}{\boldsymbol{t}}}
{
\left\lVert
\prescript{C}{P}{\boldsymbol{R}}r \prescript{P}{}{\hat{\boldsymbol{p}}} + \prescript{C}{P}{\boldsymbol{t}}
\right\rVert_2
}
\end{aligned}
\end{equation}
To support all Kalibr-compatible camera models \cite{kalibr_cam_calib} (pinhole, MEI, etc.), we adopt a general physical camera projection operator $\prescript{C}{}{\boldsymbol{\pi}}(\cdot)$ with intrinsics $\prescript{C}{}{\boldsymbol{\Xi}}$ including distortion:
\begin{align}
\prescript{C}{}{\boldsymbol{u}} &= \prescript{C}{}{\boldsymbol{\pi}}\big(\prescript{C}{}{\hat{\boldsymbol{p}}},\; \prescript{C}{}{\boldsymbol{\Xi}}\big)
\end{align}
This completes USPM's full bidirectional mapping: physical pixels $\prescript{C}{}{\boldsymbol{u}}$ forward-project to panoramic $\prescript{P}{}{\boldsymbol{u}}$ as illustrated in Fig. \ref{fig:USPM_Projection}, while $\prescript{P}{}{\boldsymbol{u}}$ inversely maps to 3D direction $\prescript{C}{}{\hat{\boldsymbol{p}}}$, then reprojects to $\prescript{C}{}{\boldsymbol{u}}$ via $\prescript{C}{}{\boldsymbol{\pi}}(\cdot)$, enabling flexible deployment across diverse multi-camera systems.
Thanks to USPM's bidirectional mapping, we achieve fast panorama construction, unified feature extraction and efficient depth initialization.
Our USPM parallelizes remapping via Parallel Unit 1 (Fig. \ref{fig:overview_pipeline}) with independent per-camera projection, reducing panorama synthesis latency and enabling real-time operation. Following PAN-SLAM \cite{pan_slam}, we extract global features on the full panorama, but adopt lightweight FAST keypoints instead of ORB to minimize computational overhead. Critically, in contrast to PAN-SLAM's depth initialization scheme, we reproject each panoramic feature back to its corresponding physical cameras and perform cross-camera triangulation to generate omnidirectional depth priors for subsequent state estimation. To enhance triangulation robustness, we implement cylindrical undistortion preprocessing from D2SLAM \cite{d2slam}, which is only applied to non-parallel camera pairs to avoid redundant computation. A dual parallel unit framework further accelerates the pipeline: Parallel Unit 2 handles adaptive undistortion and cylindrical remapping, while Parallel Unit 3 executes parallel epipolar matching \cite{svo_2_0} and depth estimation across all camera pairs.
\subsection{Hierarchical Omnidirectional Feature Alignment}
\label{subsec:hofa}
Building on SVO 2.0 \cite{svo_2_0}, we propose the HOFA framework for USPM. It adopts a task-adaptive strategy: new features undergo cross-camera depth estimation with range filtering, while pre-triangulated features directly reuse their reliable depth. Both depth sources enable panorama-level semi-direct pixel patch matching, guided by epipolar constraints from physical cameras.

For features with known 3D positions, our feature alignment based on USPM is formulated as:
\begin{equation}
\begin{aligned}
{\prescript{P}{}{\boldsymbol{u}}}^{\prime} &= 
\prescript{U}{}{\boldsymbol{\pi}} \left( \prescript{C}{B}{\boldsymbol{T}} \prescript{B_j}{B_i}{\boldsymbol{T}} \prescript{B}{C}{\boldsymbol{T}}   {\prescript{U}{}{\boldsymbol{\pi}}}^{-1}(\prescript{P}{}{\boldsymbol{u}}) \frac{1}{\prescript{C}{}{\tilde{i}}} \right) \\
{{\prescript{P}{}{\delta \boldsymbol{u}}}}^{*} &= \arg\min_{ {\prescript{P}{}{\delta \boldsymbol{u}}} } 
\sum_{\Delta \prescript{P}{}{ \boldsymbol{u} } \in \mathcal{P} } \frac{1}{2} \left\| \prescript{P}{}{I}_{j}\left( {\prescript{P}{}{\boldsymbol{u}}}^{\prime} + {{\prescript{P}{}{\delta \boldsymbol{u}}}} + \Delta \prescript{P}{}{ \boldsymbol{u} } \right) \right.  \\
& \quad \left. -  \prescript{P}{}{I}_{i}( \prescript{P}{}{\boldsymbol{u}} + \boldsymbol{A} \Delta \prescript{P}{}{ \boldsymbol{u} } ) \right\| ^{2}  \\
{\prescript{P}{}{\boldsymbol{u}}}^{\prime *} &= {\prescript{P}{}{\boldsymbol{u}}}^{\prime} + {{\prescript{P}{}{\delta \boldsymbol{u}}}}^{*} \label{eq:feature_alignment}
\end{aligned}
\end{equation}
The unified formulation in \eqref{eq:feature_alignment} defines USPM-specific panoramic alignment, using inverse depth $\prescript{C}{}{\tilde{i}}$ from physical cameras. Notations including patch $\mathcal{P}$, affine warp $\boldsymbol{A}$ and pixel correction $\prescript{P}{}{\delta \boldsymbol{u}}$ follow SVO 2.0 \cite{svo_2_0}. The first equation predicts initial feature positions in frame $j$, while the second and third equations optimize and output the refined alignment.

\begin{figure}[htbp]
    \centering
    \includegraphics[width=1\linewidth]{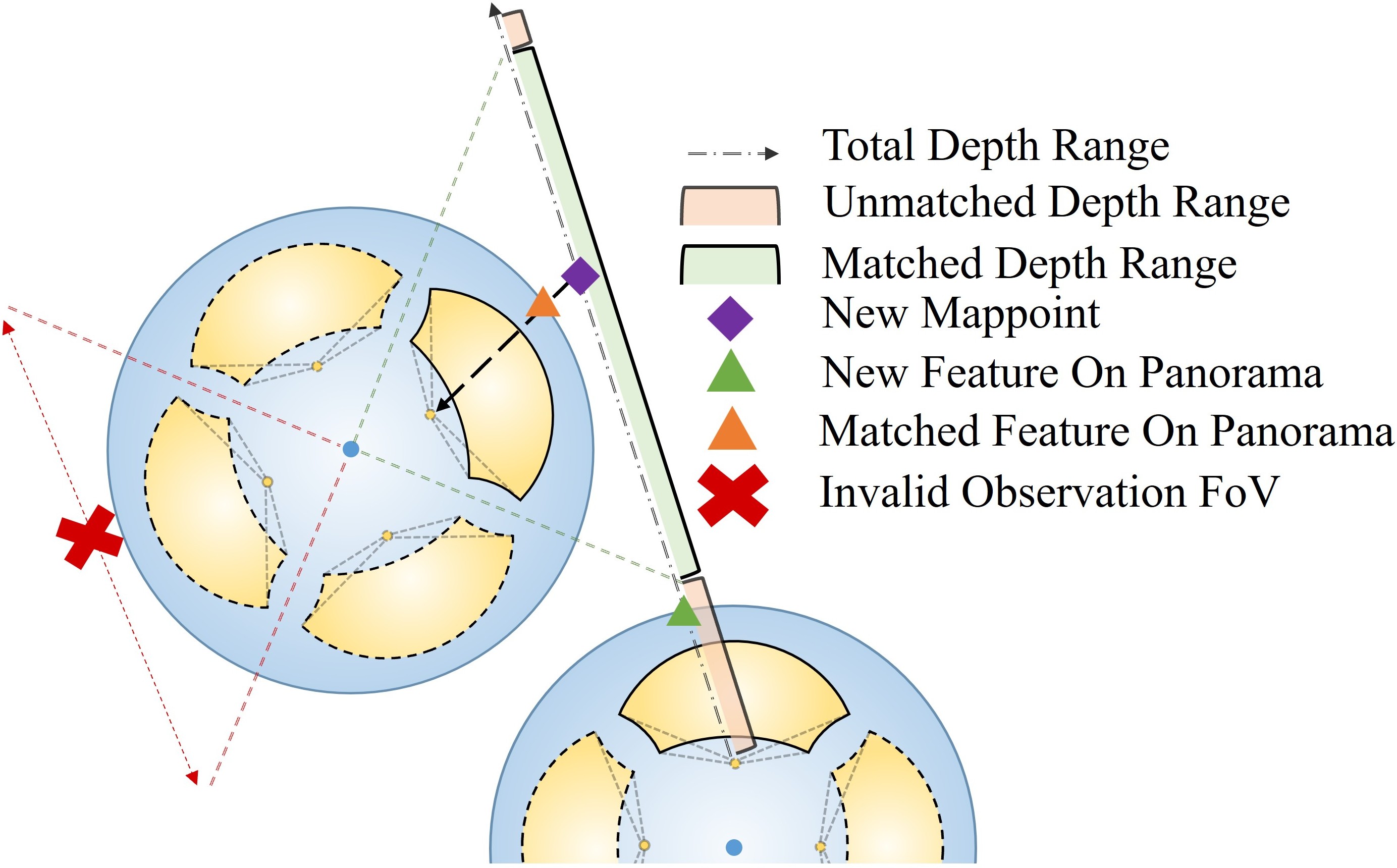}
    \caption{Geometric illustration of cross-camera inverse depth range convergence, taking the Seeker Omnidirection module(Sec. \ref{sec:selfmade_omnidirectional_results}) as an example.}
    \label{fig:adaptive_depth_Estimation_and_feature_alignment}
\end{figure}
For new features without pre-triangulated depth, HOFA performs cross-camera inverse depth estimation via the constrained photometric optimization:
\begin{align}
{\prescript{C}{}{\tilde{i}}}^{*} &= \arg\min_{ \prescript{C}{}{\tilde{i}} } 
\sum_{ \Delta \prescript{P}{}{ \boldsymbol{u} } \in \mathcal{P} } \frac{1}{2} \left\| \prescript{P}{}{I}_{j}\left( {\prescript{P}{}{\boldsymbol{u}}}^{\prime} + \Delta \prescript{P}{}{ \boldsymbol{u} } \right) \right. \notag  \\
& \quad \left. -  \prescript{P}{}{I}_{i}( \prescript{P}{}{\boldsymbol{u}} + \boldsymbol{A} \Delta \prescript{P}{}{ \boldsymbol{u} } ) \right\| ^{2}, \quad
\prescript{C}{}{\tilde{i}} \in \mathcal{I}
\end{align}
The resulting inverse depth is passed to the Vogiatzis depth filter for stable convergence, where ${\prescript{P}{}{\boldsymbol{u}}}^{\prime}$ is the panoramic projection term (Eq. \eqref{eq:feature_alignment}), and $\mathcal{I}$ is the unified feasible inverse depth set from adaptive range filtering. To derive $\mathcal{I}$, we first establish the analytical relationship between a map point's panoramic yaw angle and its inverse depth:
\begin{equation}
\begin{aligned}
    \prescript{P_{i}}{}{\tilde{\boldsymbol{v}}} &= \prescript{P_{i}}{C_{i,k}}{\tilde{\boldsymbol{R}}} \prescript{C_{i,k}}{}{\hat{\boldsymbol{p}}} \\
    \prescript{P_{i}}{}{\tilde{\boldsymbol{p}}} \prescript{C_{i,k}}{}{\tilde{i}} &= \prescript{P_{i}}{}{\tilde{\boldsymbol{v}}} + \prescript{P_{i}}{C_{i,k}}{\tilde{\boldsymbol{t}}} \prescript{C_{i,k}}{}{\tilde{i}}  \\
    \tan\left(\prescript{P_{i}}{}{\tilde{\theta}}\right) &= \frac{\prescript{P_{i}}{}{\tilde{\boldsymbol{p}}}_x}{\prescript{P_{i}}{}{\tilde{\boldsymbol{p}}}_z} = \frac{\prescript{P_{i}}{}{\tilde{\boldsymbol{v}}}_x + \prescript{P_{i}}{C_{i,k}}{\tilde{\boldsymbol{t}}}_x  \prescript{C_{i,k}}{}{\tilde{i}}}{\prescript{P_{i}}{}{\tilde{\boldsymbol{v}}}_z + \prescript{P_{i}}{C_{i,k}}{\tilde{\boldsymbol{t}}}_z \prescript{C_{i,k}}{}{\tilde{i}}} \label{eq:tan_theta}
\end{aligned}
\end{equation}
where $\prescript{P_i}{C_{i,k}}{\tilde{\boldsymbol{R}}}, \prescript{P_i}{C_{i,k}}{\tilde{\boldsymbol{t}}}$ are the rotation and translation from $k$-th camera $C_{i,k}$ to panoramic frame $P_i$ of keyframe $i$, respectively. $\tan\left(\prescript{P_{i}}{}{\tilde{\theta}}\right)$ represents the tangent of the map point’s yaw angle in the panoramic frame $P_i$.

By enforcing the map point lies within each camera's horizontal FOV ($\Theta^{\text{min}}_k < \prescript{P_i}{}{\tilde{\theta}} < \Theta^{\text{max}}_k$), we derive the constraint:
\begin{align}
\begin{cases}
    \tan\left(\prescript{P_{i}}{}{\tilde{\theta}}\right) > \min\left( \tan\left(\Theta^{\text{min}}_{k}\right), \tan\left(\Theta^{\text{max}}_{k}\right) \right) \\[4pt]
    \tan\left(\prescript{P_{i}}{}{\tilde{\theta}}\right) < \max\left( \tan\left(\Theta^{\text{min}}_{k}\right), \tan\left(\Theta^{\text{max}}_{k}\right) \right)
\end{cases}  \label{eq:fov_constraint}
\end{align}

Substituting \eqref{eq:tan_theta} into \eqref{eq:fov_constraint} solves the valid depth range for each camera. As shown in Fig. \ref{fig:adaptive_depth_Estimation_and_feature_alignment}, only cameras with FOV-epipolar line intersection are considered valid. The unified depth set is:
\begin{align}
\mathcal{I} &= \Big\{ \big[ \prescript{C_{i,k}}{}{\tilde{i}}_{\text{min}},\ \prescript{C_{i,k}}{}{\tilde{i}}_{\text{max}} \big] \,\Big\vert\, k \in \mathcal{K}_{\text{valid}} \Big\}
\end{align}
where $\mathcal{K}_{\text{valid}}$ denotes valid camera indices. This adaptive filtering significantly improves subsequent depth estimation efficiency and accuracy.
\subsection{Efficient ESKF via the Spherical Bearing Vector Residual}
\label{subsec:spherical_eskf}
We fully reuse SchurVINS' \cite{schurvins} state definition and IMU propagation pipeline for our sliding-window ESKF backend, adopting its standard 15-dimensional IMU-centric state vector:
\begin{align}\boldsymbol{\chi}_{I} &= 
\left[
\begin{array}{lllll}\prescript{G}{I}{\boldsymbol{q}}^\top & \prescript{G}{}{\boldsymbol{p}}_I^\top & \prescript{G}{}{\boldsymbol{v}}_I^\top & \boldsymbol{b}_a^\top & \boldsymbol{b}_g^\top\end{array}
\right]^\top
\end{align}
where $\prescript{G}{I}{\boldsymbol{q}}$, $\prescript{G}{}{\boldsymbol{p}}_I$, $\prescript{G}{}{\boldsymbol{v}}_I$ are IMU orientation, position and velocity in the global frame, and $\boldsymbol{b}_a$, $\boldsymbol{b}_g$ are accelerometer and gyroscope biases. IMU state propagation and covariance prediction use 4th-order Runge-Kutta integration as in SchurVINS.

\begin{figure}
    \centering
    \includegraphics[width=\linewidth]{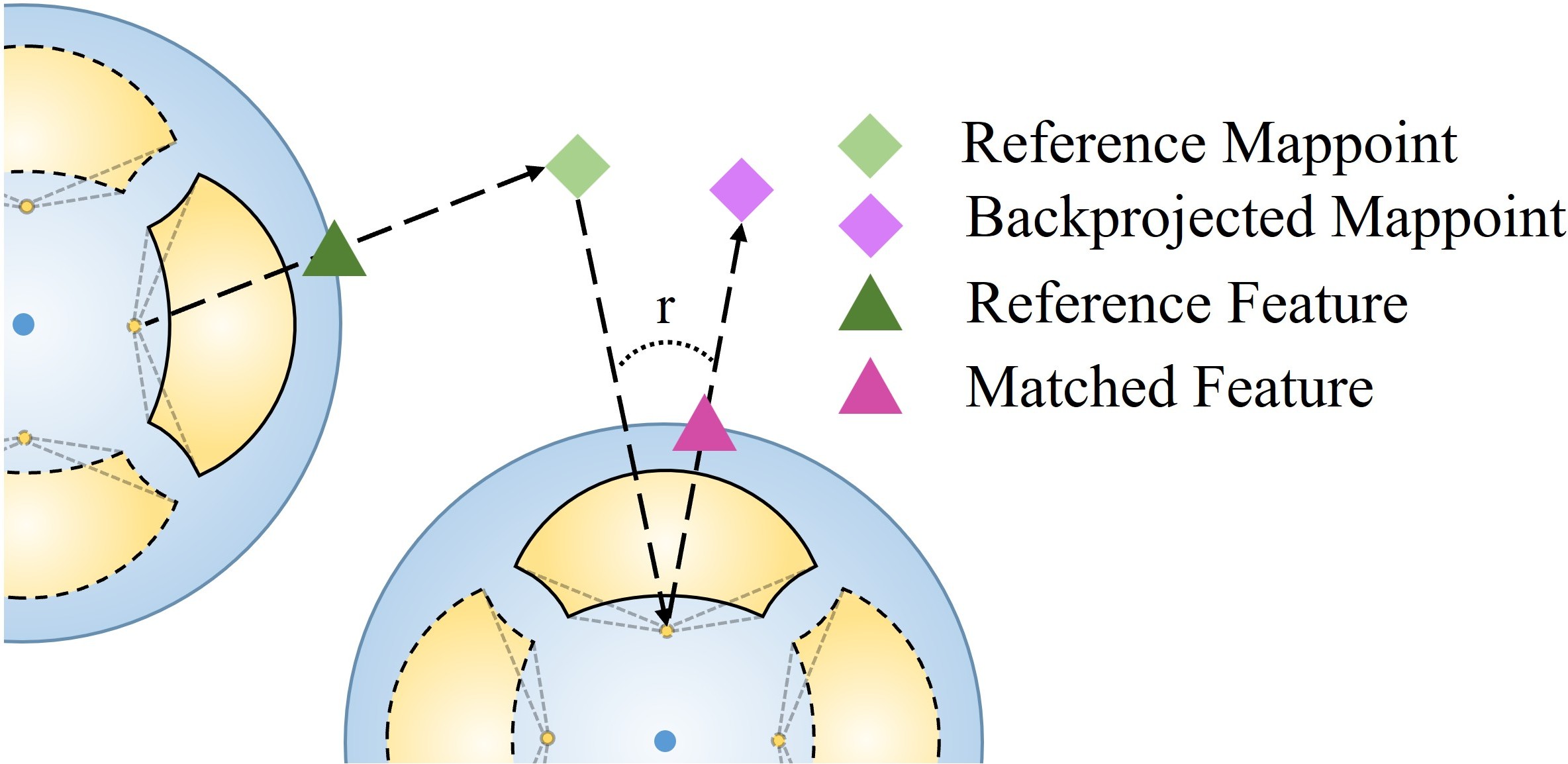}
    \caption{Geometric illustration of ESKF reprojection residual construction. $\mathbf{r}$ is the residual vector defined in \eqref{eq:vector_residual}.}
    \label{fig:spherical_residual_for_eskf}
\end{figure}
We replace SchurVINS' 2D perspective reprojection residual with a 3D spherical bearing vector residual tailored to USPM (Fig. \ref{fig:spherical_residual_for_eskf}), avoiding the strong nonlinearity of direct pixel error for omnidirectional projections \cite{360_vio}. While other panoramic SLAM systems \cite{pan_slam,rovins} also use spherical residuals, they rely on heavy graph optimization. We retain SchurVINS' full Schur complement pipeline to efficiently mitigate the additional computational overhead of our 3D residual. The spherical bearing vector residual is formulated as:
\begin{align}
\boldsymbol{r}_{i,j} &= \boldsymbol{z}_{i,j,k} - \hat{\boldsymbol{z}}_{i,j,k}, \quad
\hat{\boldsymbol{z}}_{i,j,k} = \frac{\prescript{C_{i,k}}{}{\tilde{\boldsymbol{p}}}_{j}}{\lVert \prescript{C_{i,k}}{}{\tilde{\boldsymbol{p}}}_{j} \rVert_2} \label{eq:vector_residual} 
\end{align}
where  $\boldsymbol{r}_{i,j}$ denotes the 3D spherical bearing vector residual, and observations with $\lVert \boldsymbol{r}_{i,j} \rVert_2$ exceeding a predefined threshold are discarded as outliers after the ESKF update. $\hat{\boldsymbol{z}}_{i,j,k}$ is the normalized unit sphere prediction for the $k$-th mappoint.

To perform ESKF update, we derive the residual Jacobians against system states and mappoints as detailed below:
\begin{equation}
\begin{aligned}
\boldsymbol{J}_z^r &= \boldsymbol{J}_{\delta p^W}^{p^W}
= -  \boldsymbol{I}, \quad
\boldsymbol{J}_{p^W}^{p^C} = \prescript{W}{C_{i,k}}{\tilde{\boldsymbol{R}}}^{\top}
 \\
\boldsymbol{J}_{\delta\chi^W}^{p^W} &= 
\begin{bmatrix} 
-\prescript{W}{B_i}{\tilde{\boldsymbol{R}}}[\boldsymbol{\prescript{B_i}{}{\boldsymbol{p}}_j}]_{\times}\prescript{W}{B_i}{\tilde{\boldsymbol{R}}}^{\top} & \boldsymbol{I} 
\end{bmatrix} \\
\boldsymbol{J}_{A} &= 
\boldsymbol{J}_z^r  
\boldsymbol{J}_{p^C}^z 
\boldsymbol{J}_{p^W}^{p^C}  \boldsymbol{J}_{\delta\chi^W}^{p^W} \\
\boldsymbol{J}_{\chi_{i,j}} &= \begin{bmatrix} \boldsymbol{0}_{3\times(15+6i)} & \boldsymbol{J}_{A} & \boldsymbol{0}_{3\times6(N-i-1)} \end{bmatrix} \\
\boldsymbol{J}_{p^C}^z 
&= 
\frac{\lVert \prescript{C_{i,k}}{}{\tilde{\boldsymbol{p}}} \rVert_2^2 \boldsymbol{I} - \prescript{C_{i,k}}{}{\tilde{\boldsymbol{p}}}\prescript{C_{i,k}}{}{\tilde{\boldsymbol{p}}}^{\top}}{\lVert \prescript{C_{i,k}}{}{\tilde{\boldsymbol{p}}} \rVert_2^3} \\
\boldsymbol{J}_{p_{i,j}} &= 
\boldsymbol{J}_z^r 
\boldsymbol{J}_{p^C}^z 
\boldsymbol{J}_{p^W}^{p^C} \boldsymbol{J}_{\delta p^W}^{p^W} \\
\boldsymbol{r}_{i,j} &= \boldsymbol{J}_{\chi_{i,j}} \prescript{W}{}{\delta\boldsymbol{\chi}}  + \boldsymbol{J}_{p_{i,j}}\prescript{W}{}{\delta\boldsymbol{p}}_{j} + \boldsymbol{\varepsilon}_{i,j} 
\end{aligned}
\end{equation}
where $\boldsymbol{J}_{\chi_{i,j}}$ and $\boldsymbol{J}_{p_{i,j}}$ stand for the Jacobians with respect to the system state and mappoint position respectively, $\boldsymbol{J}_{p^C}^z$, $\boldsymbol{J}_{p^W}^{p^C}$, $\boldsymbol{J}_{\delta\chi^W}^{p^W}$, $\boldsymbol{J}_z^r$, and $\boldsymbol{J}_{\delta p^W}^{p^W}$ are intermediate Jacobians in the chain rule. $\boldsymbol{\varepsilon}_{i,j}$ is the measurement noise, $\boldsymbol{I}$ is the identity matrix, and $\left[\cdot\right]_{\times}$ denotes the skew-symmetric matrix operator.

Adopting the noise model from SchurVINS \cite{schurvins}, we stack all residuals and Jacobians to form the global linearized system:
\begin{equation}
\boldsymbol{r} = \begin{bmatrix} \boldsymbol{J}_\chi & \boldsymbol{J}_p \end{bmatrix} \begin{bmatrix} \prescript{W}{}{\delta\boldsymbol{\chi}} \\ \prescript{W}{}{\delta\boldsymbol{p}} \end{bmatrix} + \boldsymbol{n}
\end{equation}
where $\boldsymbol{n}$ is the stacked measurement noise, $\boldsymbol{J}_\chi$ and $\boldsymbol{J}_p$ are the stacked state and mappoint Jacobians, respectively. We directly apply the Schur complement-based ESKF update from \cite{schurvins} to avoid high-dimensional joint optimization:
\begin{equation}
\begin{aligned}
\boldsymbol{K} &= \boldsymbol{P} \boldsymbol{J}_{\chi'}^\top \left( \boldsymbol{J}_{\chi'} \boldsymbol{P} \boldsymbol{J}_{\chi'}^\top + \boldsymbol{R}_{\chi'} \right)^{-1} \\
\boldsymbol{\Delta \chi} &= \boldsymbol{K} \boldsymbol{r}_{\chi'} \\
\boldsymbol{P} &= \left( \boldsymbol{I} - \boldsymbol{K} \boldsymbol{J}_{\chi'} \right) \boldsymbol{P} \left( \boldsymbol{I} - \boldsymbol{K} \boldsymbol{J}_{\chi'} \right)^\top + \boldsymbol{K} \boldsymbol{R}_{\chi'} \boldsymbol{K}^\top \\
\boldsymbol{\chi} &= \boldsymbol{\chi} \oplus \boldsymbol{\Delta \chi}
\end{aligned}
\end{equation}
where $\boldsymbol{r}_{\chi'}$, $\boldsymbol{J}_{\chi'}$, and $\boldsymbol{R}_{\chi'}$ are the equivalent residual, Jacobian and covariance from Schur complement. Mappoints are updated independently after state refinement.
\section{EXPERIMENTAL RESULTS}

This section presents comprehensive evaluations of Sphere-VIO. We first describe the experimental hardware platform in Sec. \ref{sec:exp01_setup}, then compare with state-of-the-art baselines on public datasets in Sec. \ref{sec:exp02_public_dataset}. We further conduct ablation studies on our custom omnidirectional dataset and across diverse camera configurations in Sec. \ref{sec:exp03_ablation}. Finally, we analyze the system's computational efficiency in in Sec. \ref{sec:exp04_efficiency}.
\subsection{Experiments Setup} \label{sec:exp01_setup}
\noindent\textbf{Datasets and Metrics}. We evaluate the proposed method on four public and custom datasets, including EuRoC \cite{euroc_dataset}, TUM-VI \cite{tumvi_dataset}, HILTI 2022 \cite{hilti_dataset}, and our self-collected omnidirectional multi-camera dataset. Two quantitative metrics are adopted for evaluation: the SE(3)-aligned ATE RMSE calculated via the evo tool in meters, and the average per-frame processing time in seconds.

\noindent\textbf{Baselines}. We compare our approach against four state-of-the-art visual-inertial SLAM systems (MAVIS \cite{mavis}, ORB-SLAM3 \cite{orb_slam3}, VINS-Fusion \cite{qin2018vins}, SchurVINS \cite{schurvins}), with all inherent loop-closure modules disabled for a fair evaluation.


\noindent\textbf{Implement Details}. Unless otherwise stated, all experiments are performed on a Dell G15 5530 Laptop (Intel Core i7-13650HX, 32 GB DDR5, Ubuntu 20.04 with ROS Noetic) in pure CPU mode. Additional lightweight tests on an Intel NUC 13 embedded platform (16 GB memory)  are marked.
\subsection{Comparison With Baselines on Public Datasets} \label{sec:exp02_public_dataset}

\begin{figure}[!t]
    \centering
    \begin{minipage}{0.4\linewidth}
        \centering
        \includegraphics[width=\linewidth]{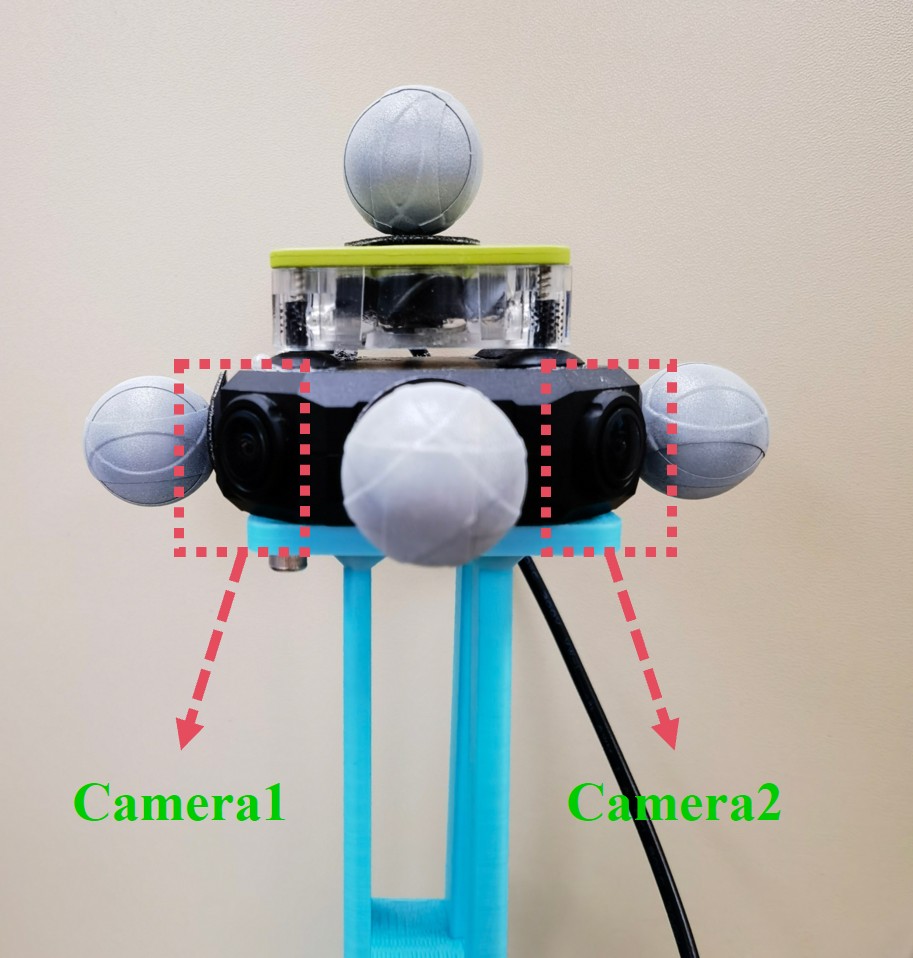}
        \par\small (a) Front view
    \end{minipage}
    \hspace{1em}
    \begin{minipage}{0.4\linewidth}
        \centering
        \includegraphics[width=\linewidth]{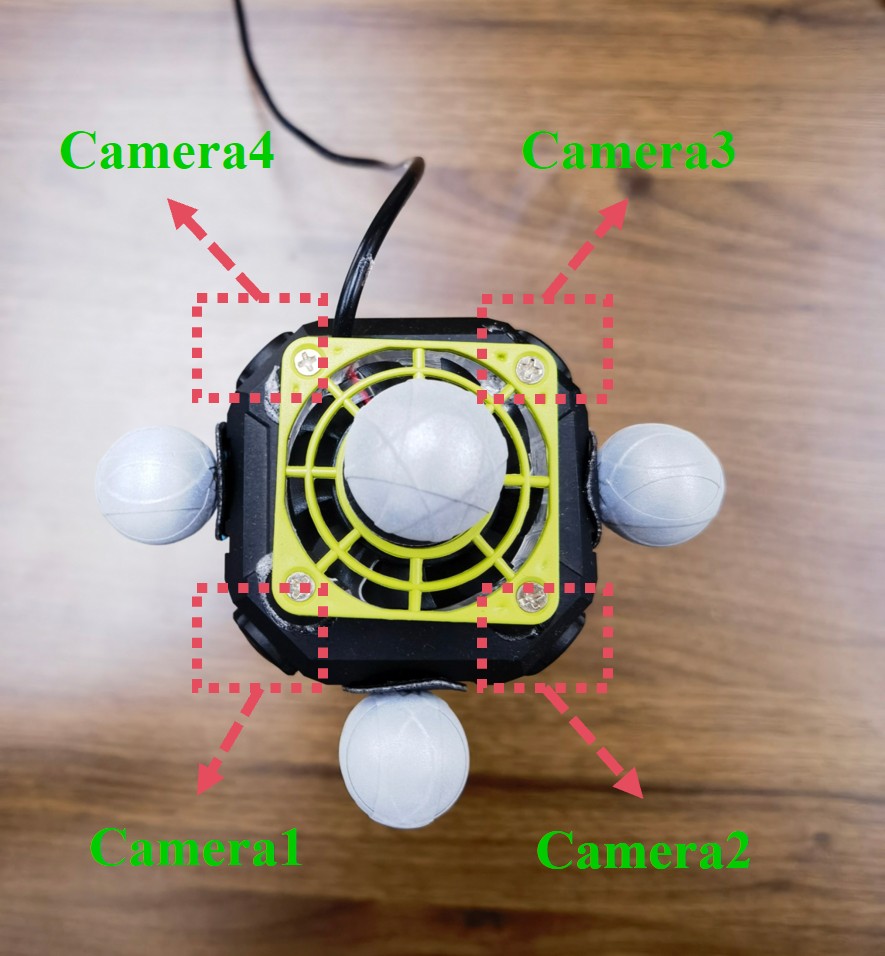}
        \par\small (b) Top view
    \end{minipage}
    \caption{Seeker OMNI-D omnidirectional multi-camera rig. Four \ang[round-mode=none]{206} fisheye cameras + industrial IMU for omnidirectional perception.}
    \label{fig:seeker_omni_d_photo}
\end{figure}

\begin{figure*}[htbp]
    \centering
    \begin{subfigure}{0.24\textwidth}
        \centering
        \includegraphics[width=\linewidth, trim=15 10 15 10, clip]{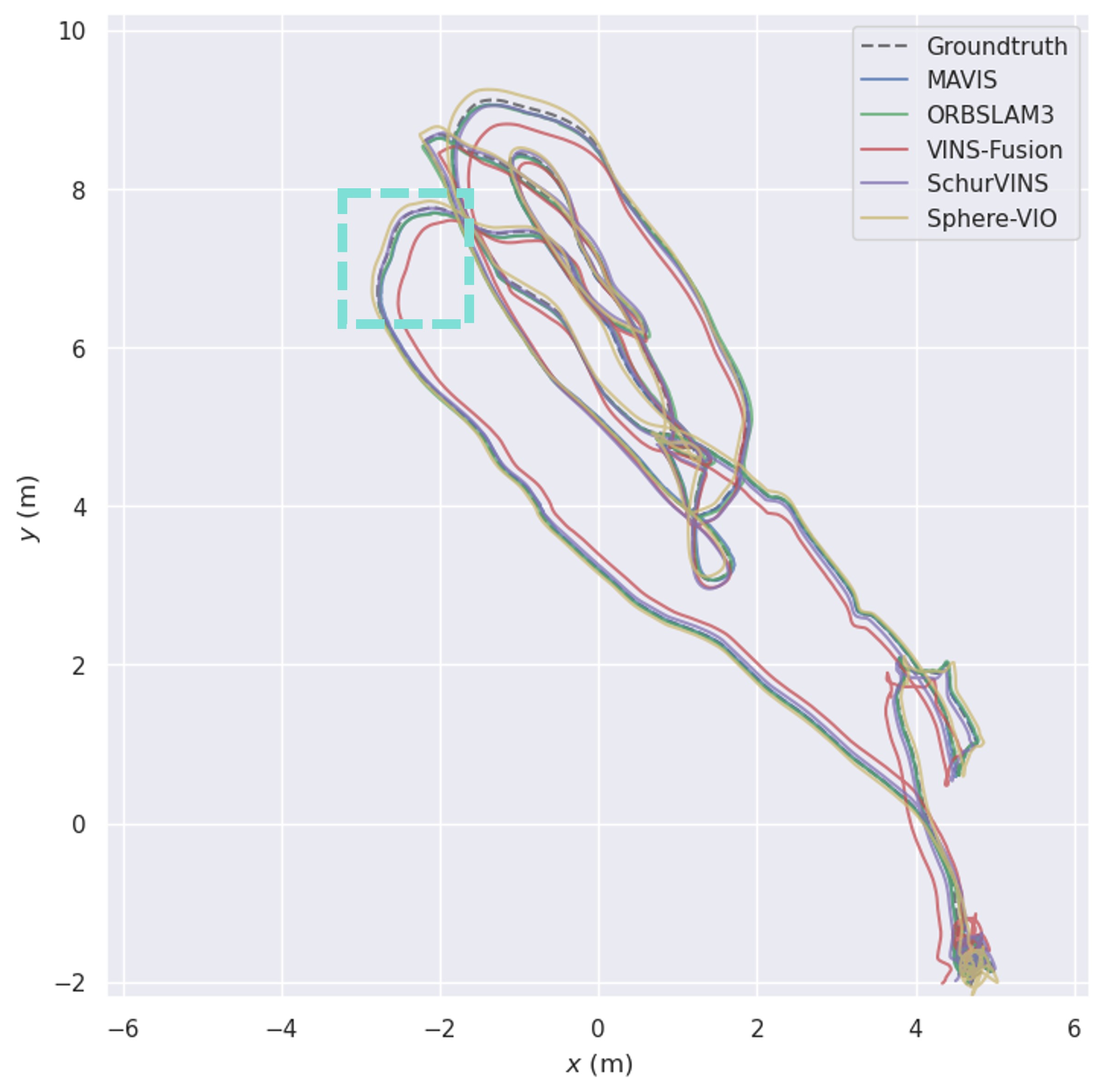}
    \end{subfigure}
    \hfill
    \begin{subfigure}{0.24\textwidth}
        \centering
        \includegraphics[width=\linewidth, trim=15 10 15 10, clip]{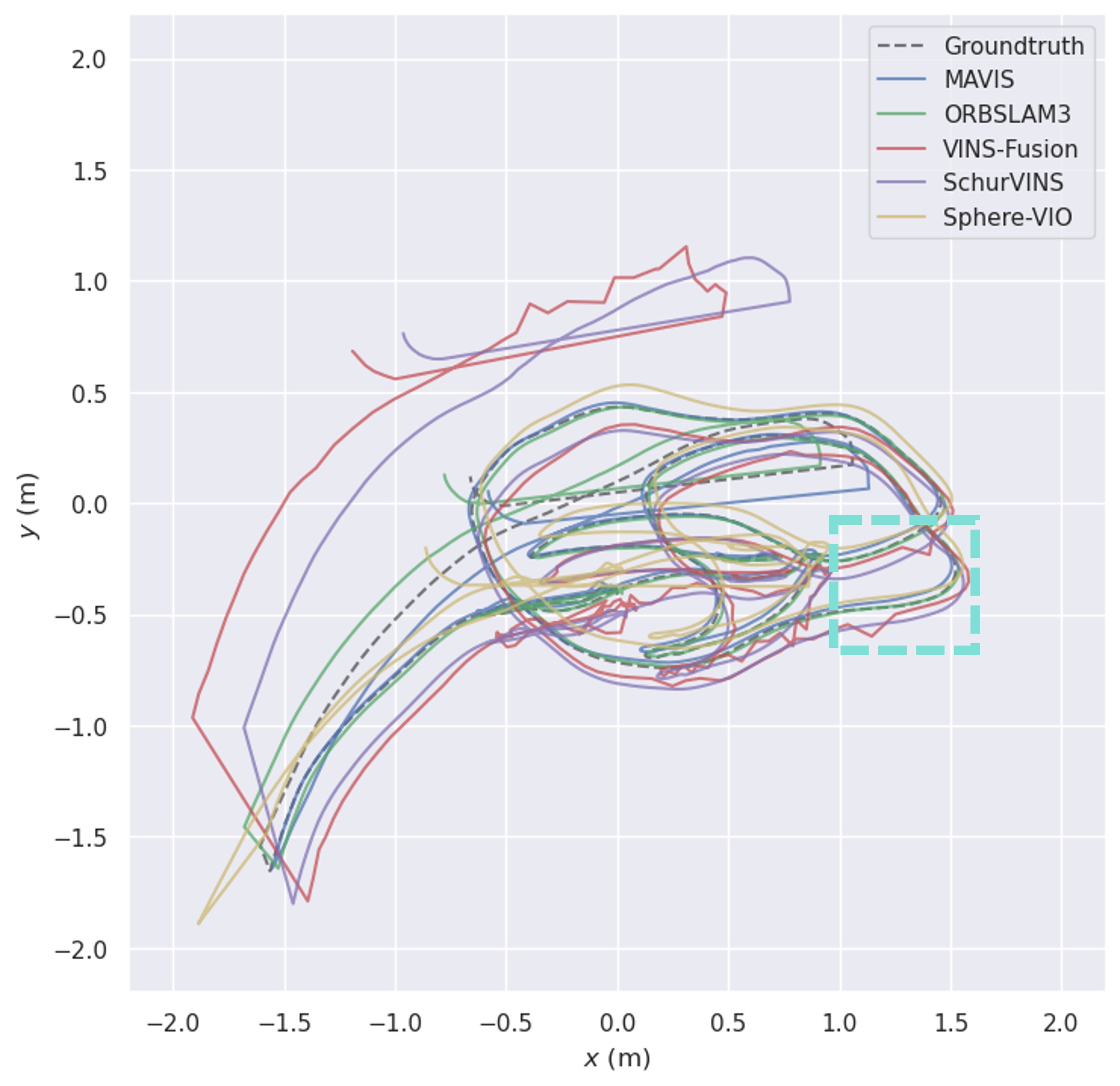}
    \end{subfigure}
    \hfill
    \begin{subfigure}{0.24\textwidth}
        \centering
        \includegraphics[width=\linewidth, trim=15 10 15 10, clip]{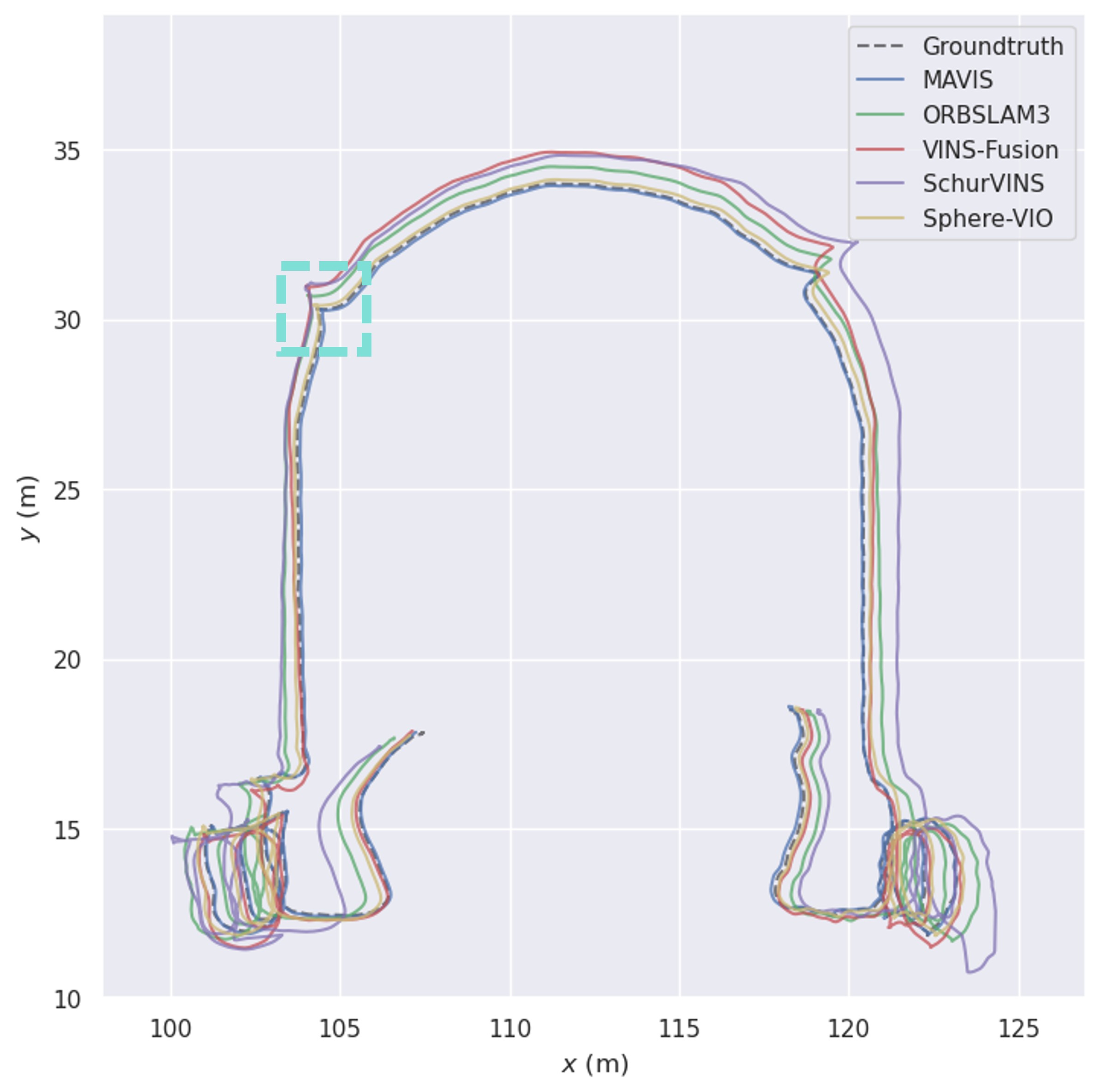}
    \end{subfigure}
    \hfill
    \begin{subfigure}{0.24\textwidth}
        \centering
        \includegraphics[width=\linewidth, trim=15 10 15 10, clip]{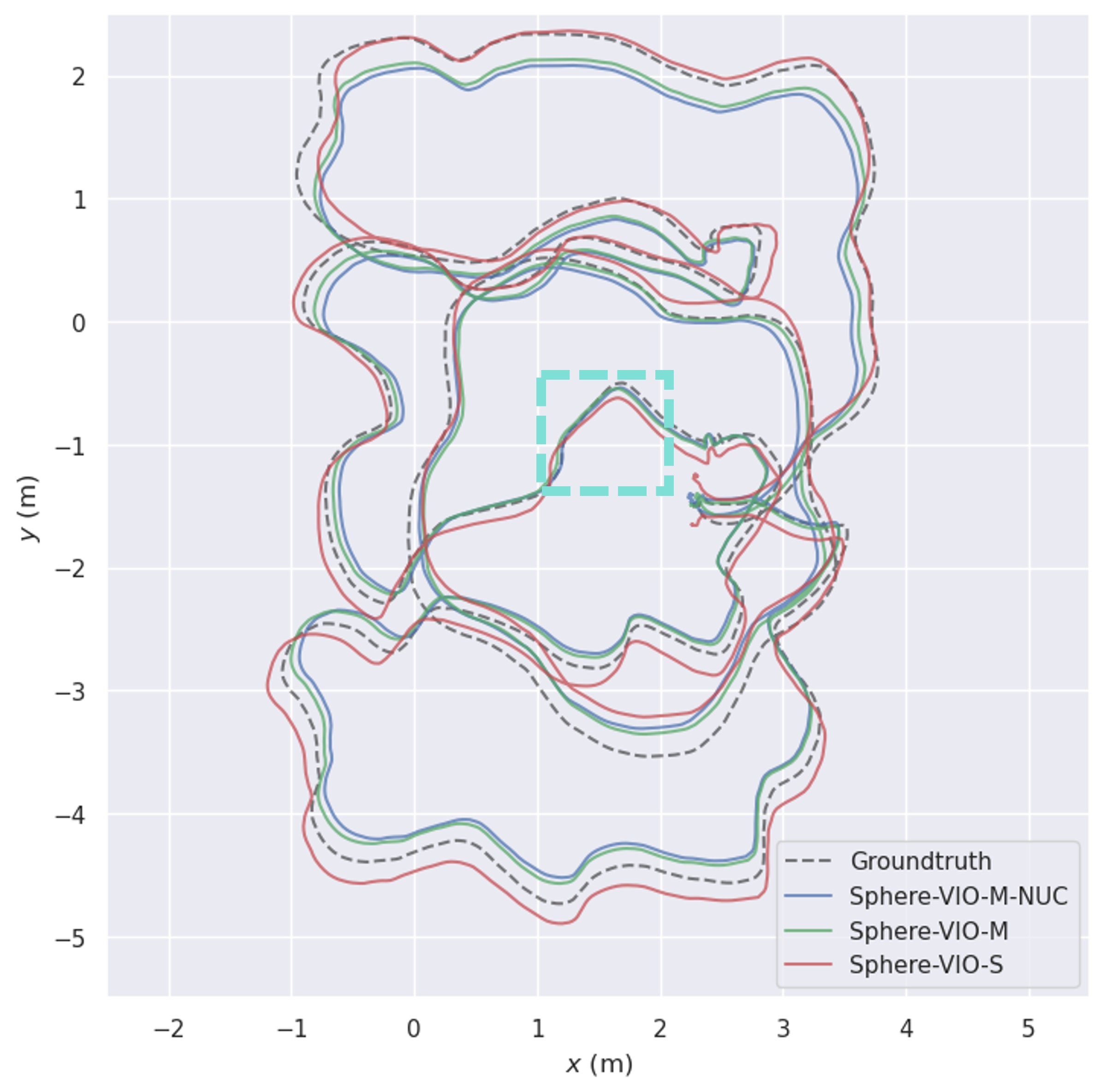}
    \end{subfigure}

    \vspace{-0.2em} 
    
    \begin{minipage}{0.24\textwidth}
        \centering
        \includegraphics[width=\linewidth]{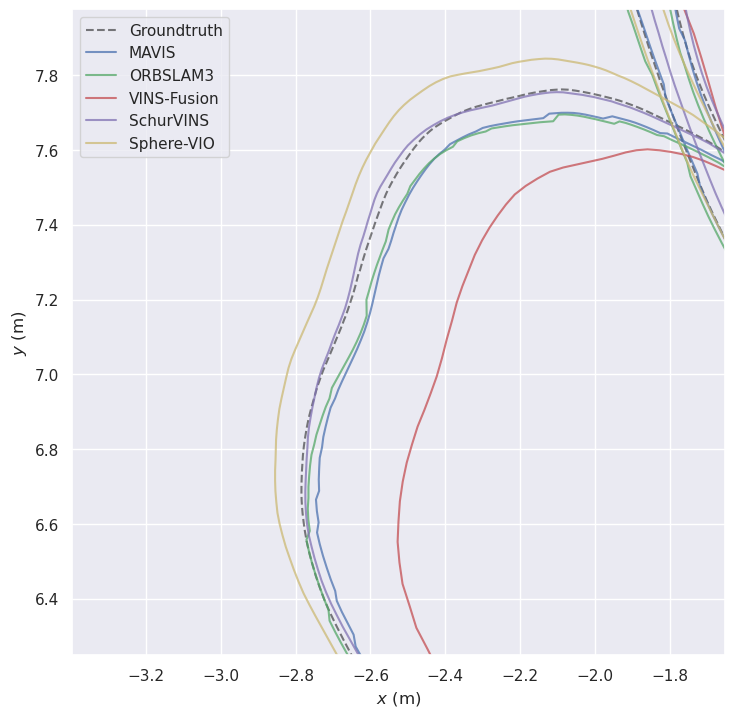}
        \vspace{0.1em}
    \end{minipage}
    \hfill
    \begin{minipage}{0.24\textwidth}
        \centering
        \includegraphics[width=\linewidth]{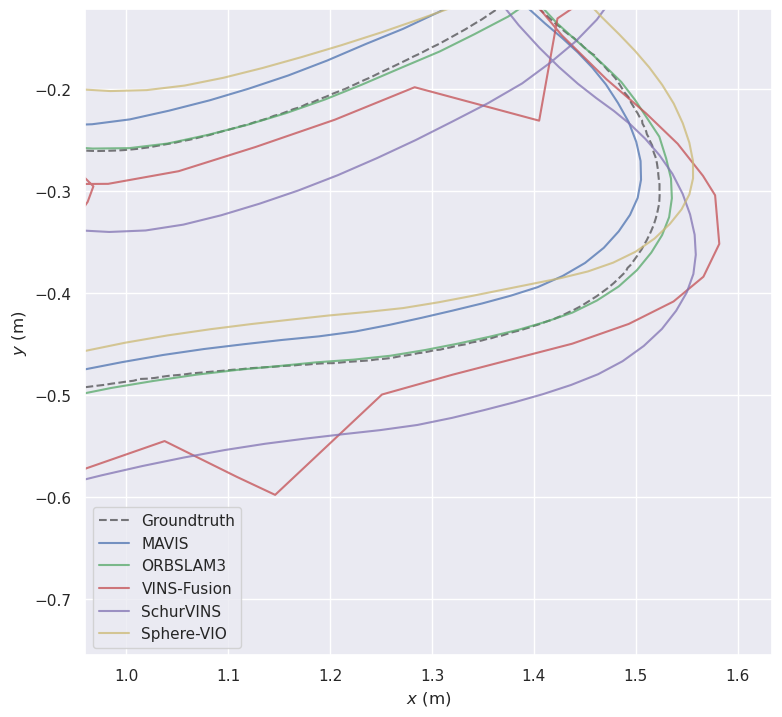}
        \vspace{0.1em}
    \end{minipage}
    \hfill
    \begin{minipage}{0.24\textwidth}
        \centering
        \includegraphics[width=\linewidth]{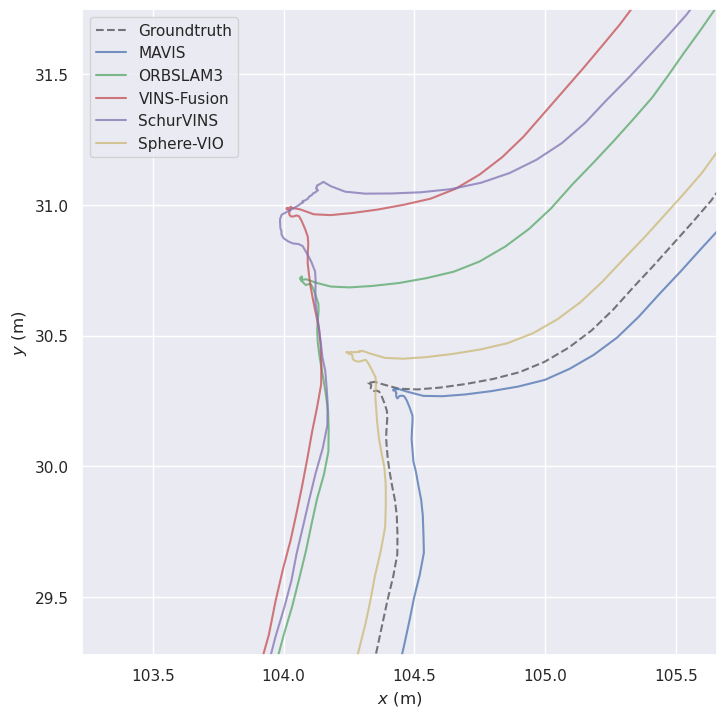}
        \vspace{0.1em}
    \end{minipage}
    \hfill
    \begin{minipage}{0.24\textwidth}
        \centering
        \includegraphics[width=\linewidth]{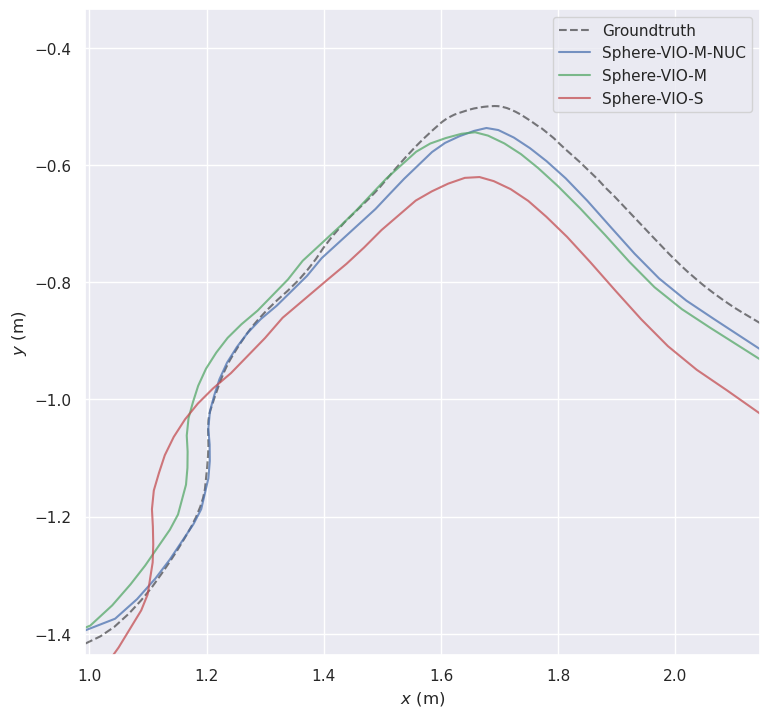}
        \vspace{0.1em}
    \end{minipage}


    \vspace{-\baselineskip} 
    \setlength{\abovecaptionskip}{0pt} 
    \vspace{-2pt} 
    
    \caption{Trajectory comparisons across four datasets. The first row shows the full trajectories corresponding to the available ground truth segments, while the second row displays the enlarged views within the corresponding blue boxes. From left to right: EuRoC MH-01, TUM-VI Corridor 5, HILTI 2022 Exp 16, and our custom omnidirectional dataset Sequence 01.}
    \label{fig:trajectory_viewer}
\end{figure*}

\begin{table*}[!t]
\centering
\caption{ATE RMSE (m) on EuRoc datasets \cite{euroc_dataset}, SE(3)-aligned. Results per sequence are colored: \protect\boxfirst{best}, \protect\boxsecond{second}, \protect\boxthird{third}. Camera(Cm): S= Stereo, M= Multi-Camera. Backend(Bk): G= Graph-based , F= Filter-based. SV: our proposed Sphere-VIO. All baseline results are manually evaluated by the authors.}
\label{tab:euroc_result}
\setlength{\tabcolsep}{1.0mm} 
\begin{tabular}{l|c|c|ccccc|ccccc|c}
\toprule[1.0pt] 
\multirow{1}{*}{Method} 
& Bk & Cm
& MH-01 & MH-02 & MH-03 & MH-04 & MH-05 & V-101 & V-102 & V-103 & V-201 & V-202 & \multirow{1}{*}{Avg} \\
\midrule[0.6pt]
ORB-SLAM3\tnote{1}\,\,
& G & S & \colorsecond\num{0.046072} & \colorfirst\num{0.030435} & \colorsecond\num{0.031254} & \colorfirst\num{0.049507} & \colorsecond\num{0.08347} & \colorfirst\num{0.03808} & \colorfirst\num{0.02016} & \colorfirst\num{0.043871} & \colorfirst\num{0.032611} & \colorfirst\num{0.024461} & \colorfirst\num{0.0399921} \\
MAVIS\tnote{1}\,\, 
& G & S & \colorfirst\num{0.043019} & \colorsecond\num{0.034437} & \colorfirst\num{0.027566} & \colorsecond\num{0.067343} & \colorfirst\num{0.073927} & \colorsecond\num{0.041982} &  \num{0.050592} & \colorsecond\num{0.056926} & \colorthird\num{0.048346} & \colorsecond\num{0.031173} & \colorsecond\num{0.0475311}   \\
VINS-Fusion\tnote{1}\,\, 
& G & S & \num{0.252745} & \num{0.21694} & \num{0.325944} & \num{0.412998} & \num{0.307328} & \num{0.113173} & \num{0.107724} & \num{0.11356} & \num{0.09949} & \num{0.123933} & \num{0.2073835} \\
\midrule[0.3pt]
SchurVINS\tnote{1}\,\,
& F & S & \colorthird\num{0.073485} & \colorthird\num{0.067382} & \colorthird\num{0.07406} & \colorthird\num{0.145599} & \colorthird\num{0.135258} & \colorthird\num{0.050109} & \colorthird\num{0.040831} & \num{0.072644} & \colorsecond\num{0.044586} & \colorthird\num{0.075476} & \colorthird\num{0.077943} \\
SV (ours)\tnote{2}\,\,
& F & S & \num{0.099552} & \num{0.098017} & \num{0.15243} & \num{0.178848} & \num{0.219177} & \num{0.055978} & \colorsecond\num{0.038596} & \colorthird\num{0.065212} & \num{0.065223} & \num{0.099195} & \num{0.1072228} \\
\bottomrule[1.0pt]
\end{tabular}
\end{table*}
\begin{table*}[!htbp]
\centering
\caption{ATE RMSE (m) on TUM-VI datasets \cite{tumvi_dataset}. c*: Corridor*; r*: Room*. Other evaluation details as in Table \ref{tab:euroc_result}.}
\label{tab:tumvi_result}
\setlength{\tabcolsep}{1.2mm} 
\begin{tabular}{l|c|c|ccccc|cccccc|c}
\toprule[1.0pt] 
\multirow{1}{*}{Method} 
& Bk & Cm
& c1 & c2 & c3 & c4 & c5 & r1 & r2 & r3 & r4 & r5 & r6 & \multirow{1}{*}{Avg} \\
\midrule[0.6pt] 
ORB-SLAM3\tnote{1}\,\, 
& G & S
& \colorfirst\num{0.152402} & \colorfirst\num{0.092776} & \colorthird\num{0.3775} & \colorsecond\num{0.085495} & \colorfirst\num{0.075133} & \colorsecond\num{0.015688} & \colorsecond\num{0.017524} & \colorsecond\num{0.031369} & \colorsecond\num{0.013381} & \colorsecond\num{0.010682} & \colorsecond\num{0.009617} & \colorsecond\num{0.080142455} \\
MAVIS\tnote{1}\,\, 
& G & S 
& \colorsecond\num{0.172349} & \colorthird\num{0.148681} & \colorfirst\num{0.299883} & \colorfirst\num{0.069353} & \colorsecond\num{0.080944} & \colorfirst\num{0.012017} & \colorfirst\num{0.008773} & \colorfirst\num{0.025688} & \colorfirst\num{0.012129} & \colorfirst\num{0.008391} & \colorfirst\num{0.007238} & \colorfirst\num{0.076916273} \\
VINS-Fusion\tnote{1}\,\, 
& G & S 
& \num{0.862748} & \num{1.068069} & \num{1.437794} & \num{0.248192} & \num{0.434693} & \num{0.081474} & \colorthird\num{0.057642} & \num{0.113015} & \num{0.052567} & \num{0.185524} & \num{0.04632} & \num{0.417094364} \\
\midrule[0.3pt] 
SchurVINS\tnote{1}\,\,
& F & S 
& \num{0.933659} & \num{0.896589} & \num{0.668317} & \colorthird\num{0.118412} & \num{0.319754} & \num{0.050267} & \num{0.093382} & \colorthird\num{0.049212} & \num{0.044177} & \colorthird\num{0.064927} & \colorthird\num{0.022897} & \num{0.296508455} \\
SV (ours)\tnote{2}\,\,
& F & S
& \colorthird\num{0.247231} & \colorsecond\num{0.134798} & \colorsecond\num{0.372774} & \num{0.191993} & \colorthird\num{0.212627} & \colorthird\num{0.042957} & \num{0.189778} & \num{0.117596} & \colorthird\num{0.030842} & \num{0.066056} & \num{0.063435} & \colorthird\num{0.151826091} \\
\bottomrule[1.0pt] 
\end{tabular}
\end{table*}
We conduct evaluations sequentially on three public datasets with distinct challenges: the pinhole-based EuRoC MAV dataset featuring low texture and aggressive fast motion (Table \ref{tab:euroc_result}), the challenging TUM-VI indoor dataset with large-range motion and variable illumination (Table \ref{tab:tumvi_result}), and the HILTI 2022 dataset characterized by sparse textures, intensive movement and uneven lighting (Table \ref{tab:hilti2022_result}). Across all benchmarks,  MAVIS obtain overall top-tier localization accuracy. Our filter-driven Sphere-VIO yields an average ATE RMSE of 0.1072 m on EuRoC, surpassing VINS-Fusion by 48.3\% but trailing SchurVINS, as our spherical bearing residual targets wide-FOV multi-camera setups rather than pinhole scenarios optimized by conventional planar reprojection residuals. On the TUM-VI dataset, our method obtains an ATE RMSE of 0.1518 m. It outperforms the filter-based SchurVINS by 48.8\% and surpasses the graph-optimized VINS-Fusion by 63.6\%. Notably, our framework operates without graph optimization or global mapping, and our unified spherical formulation brings prominent robustness for long-corridor environments where stereo-based alternatives suffer drastic drift. On HILTI 2022, Sphere-VIO ranks second overall with 0.1821 m ATE RMSE only inferior to MAVIS, outperforming VINS-Fusion, ORB-SLAM3 and SchurVINS by 35.3\%, 39.7\% and 55.7\% respectively, as multi-camera spherical modeling ensures stable tracking under harsh conditions while narrow-FOV stereo pipelines suffer severe drift. To guarantee fair evaluation throughout all tests, VINS-Fusion adopts its official EuRoC-derived configuration with one keyframe per two input frames and backend optimization restricted to keyframes only. It retains identical settings on TUM-VI and HILTI 2022.
\begin{figure*}[ht]
    \centering
    \setlength{\tabcolsep}{6pt} 
    
    \begin{tikzpicture}[remember picture, every node/.style={inner sep=0pt, outer sep=0pt}]
        \def\l_title_x{25em} 
        
        \node (camera_block) {
            \begin{tabular}{c *{4}{c}}
                & \textbf{Seeker Omnidirection} 
                & \textbf{HILTI 2022} 
                & \textbf{Seeker Stereo} 
                & \textbf{TUM-VI Stereo} \\
                \noalign{\vspace{0.3em}}
                
                & \raisebox{-0.5\height}{\includegraphics[width=0.2\textwidth]{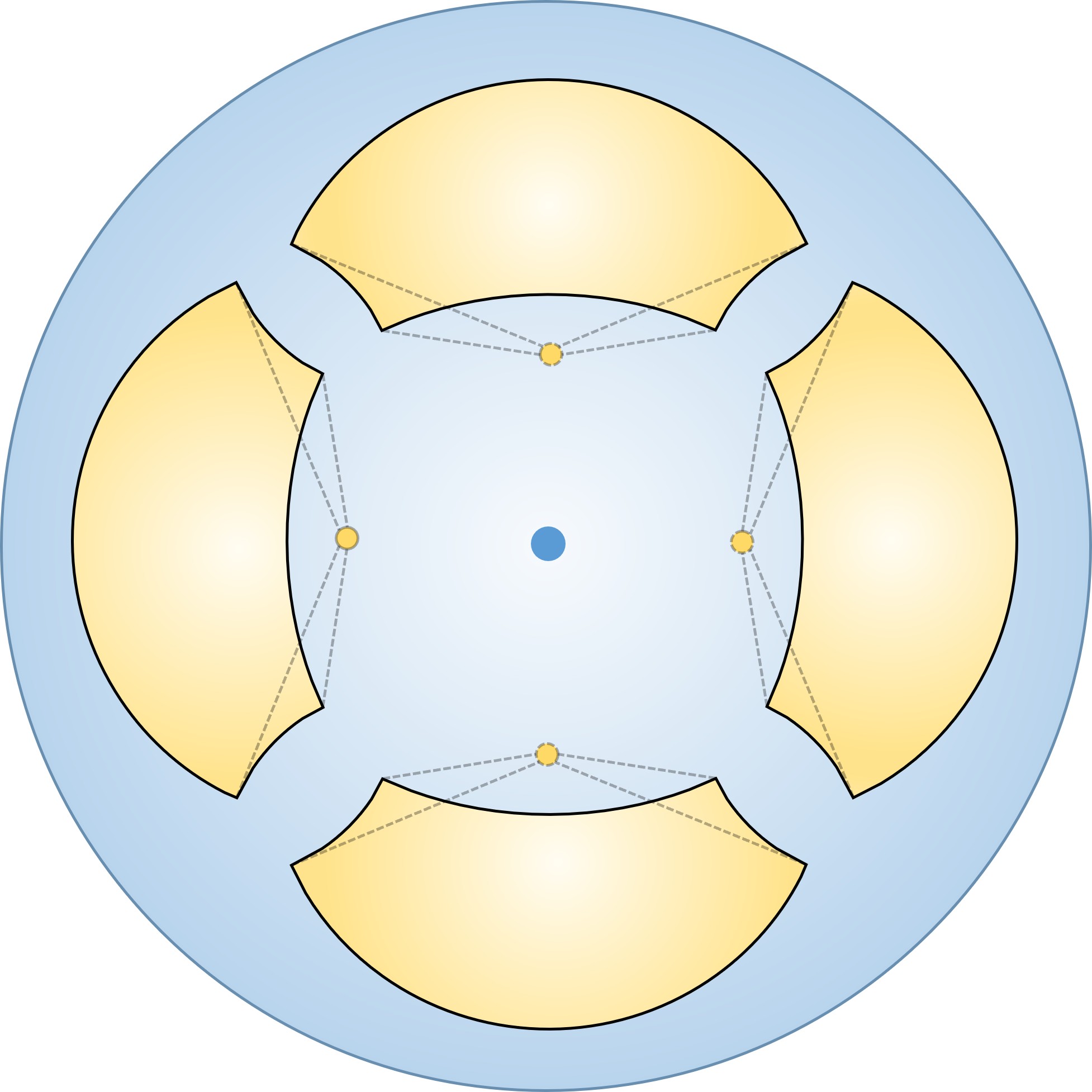}}
                & \raisebox{-0.5\height}{\includegraphics[width=0.2\textwidth]{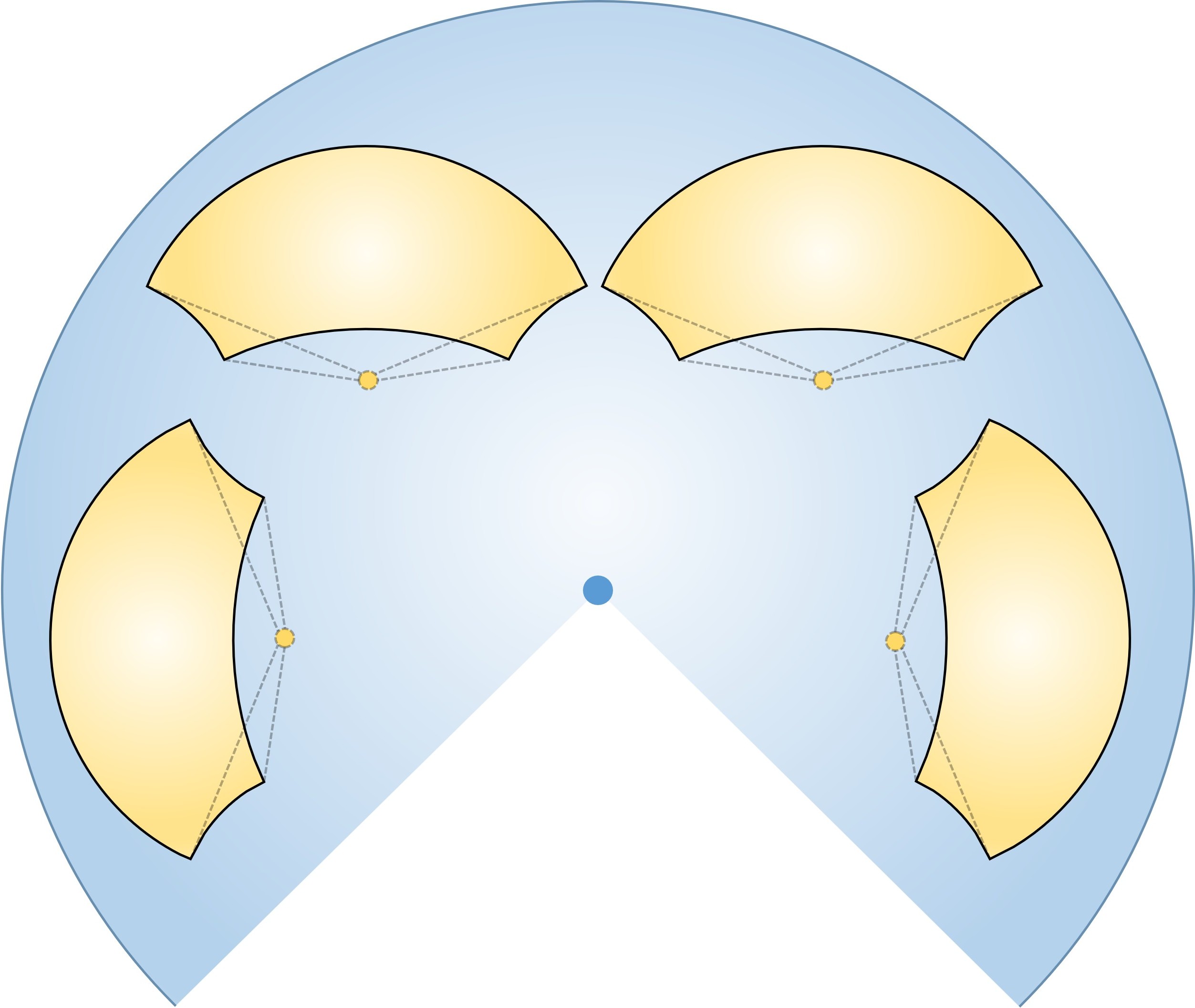}}
                & \raisebox{-0.5\height}{\includegraphics[width=0.2\textwidth]{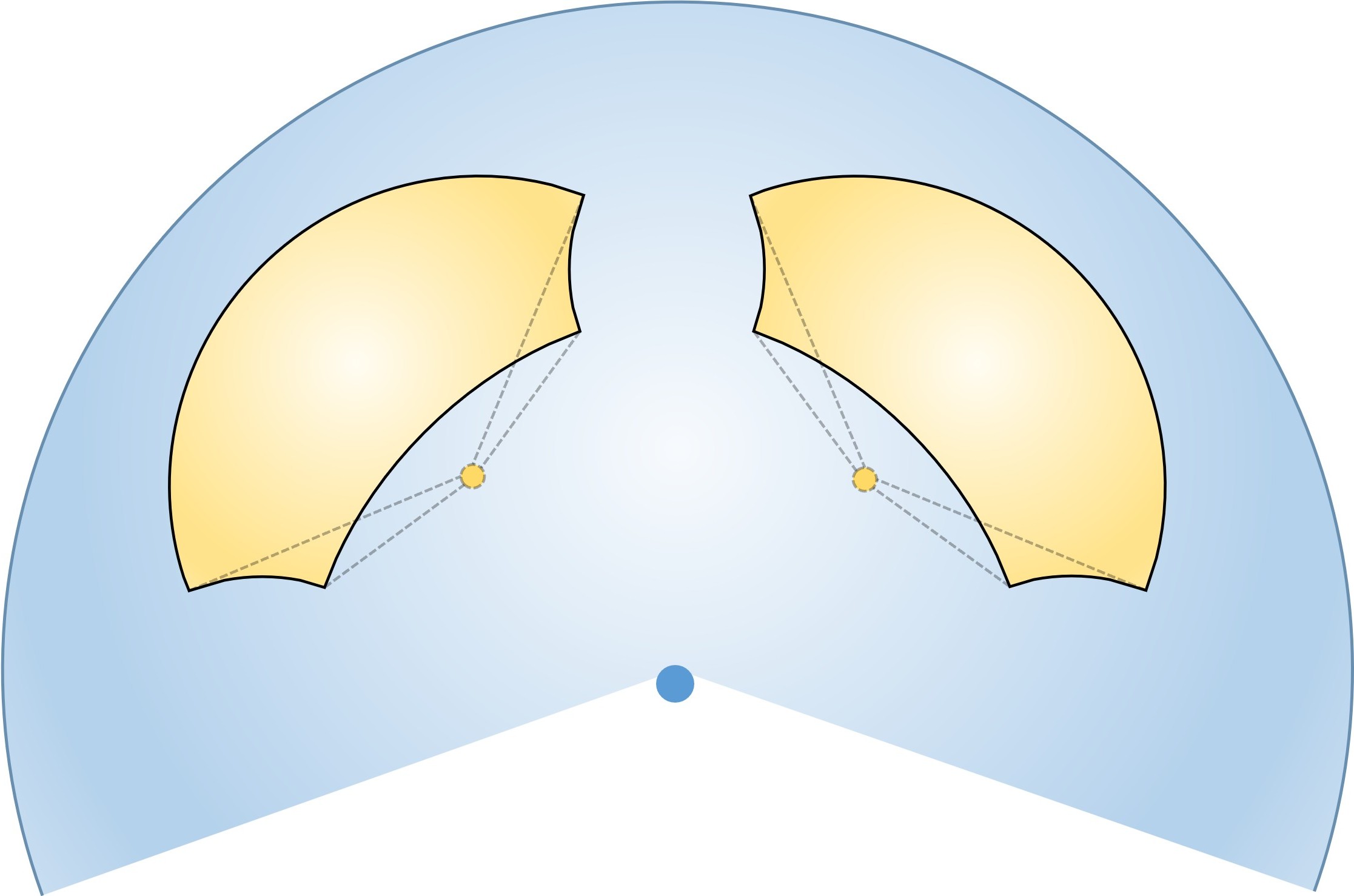}}
                & \raisebox{-0.5\height}{\includegraphics[width=0.2\textwidth]{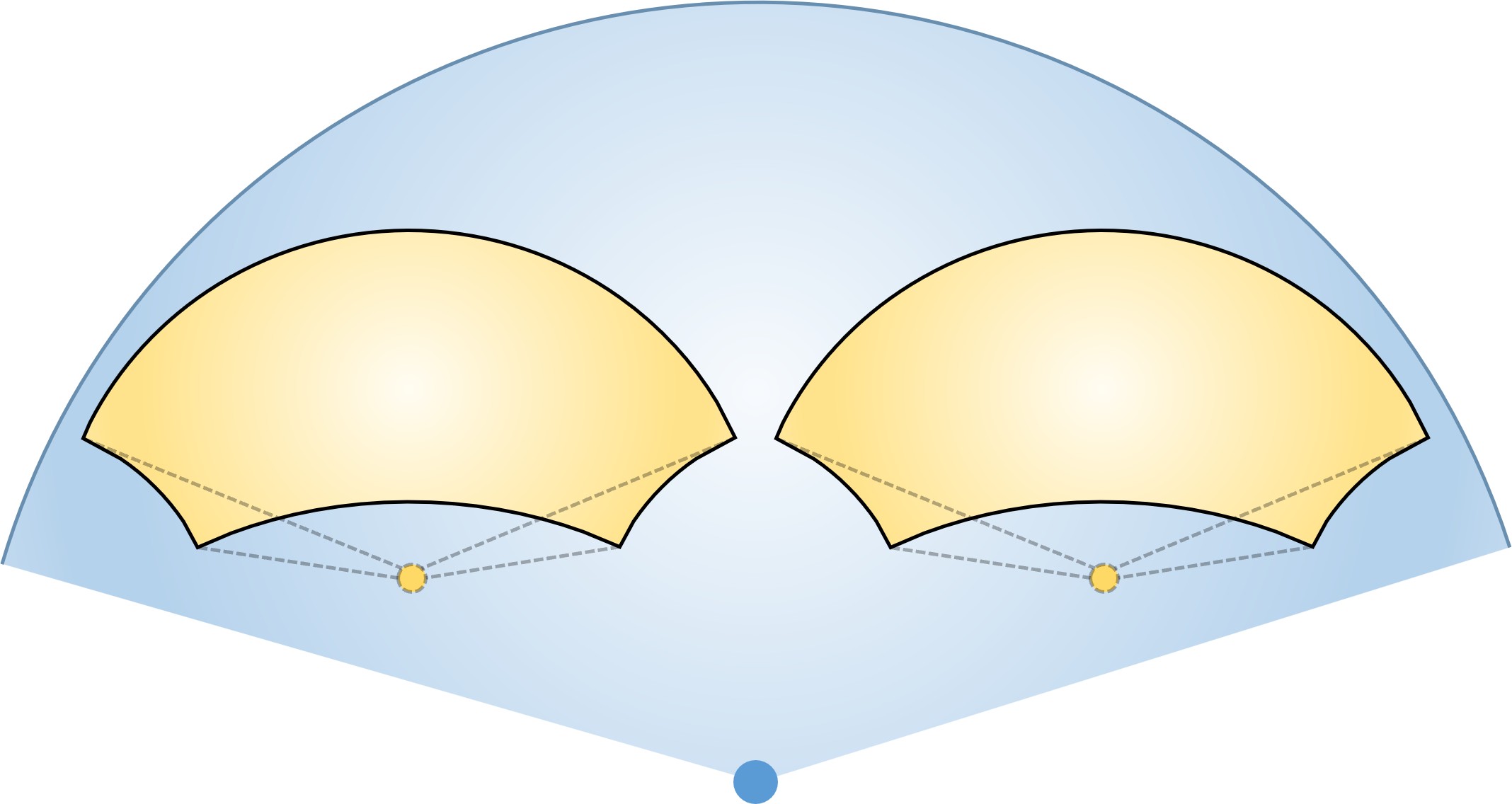}} \\
            \end{tabular}
        };
        
        \node[
            rotate=90, 
            anchor=center, 
            font=\bfseries,
            xshift=0em, 
            yshift=\l_title_x
        ] at ( {camera_block.center}) {Camera Configuration};
        
        \node[yshift=-0.3em, anchor=north] (triangulation_block) at ({camera_block.south}) {
            \begin{minipage}{0.92\textwidth}
                \centering
                \begin{tikzpicture}[inner sep=0pt, outer sep=0pt]
                    \node (img1) {\includegraphics[height=0.19\textheight]{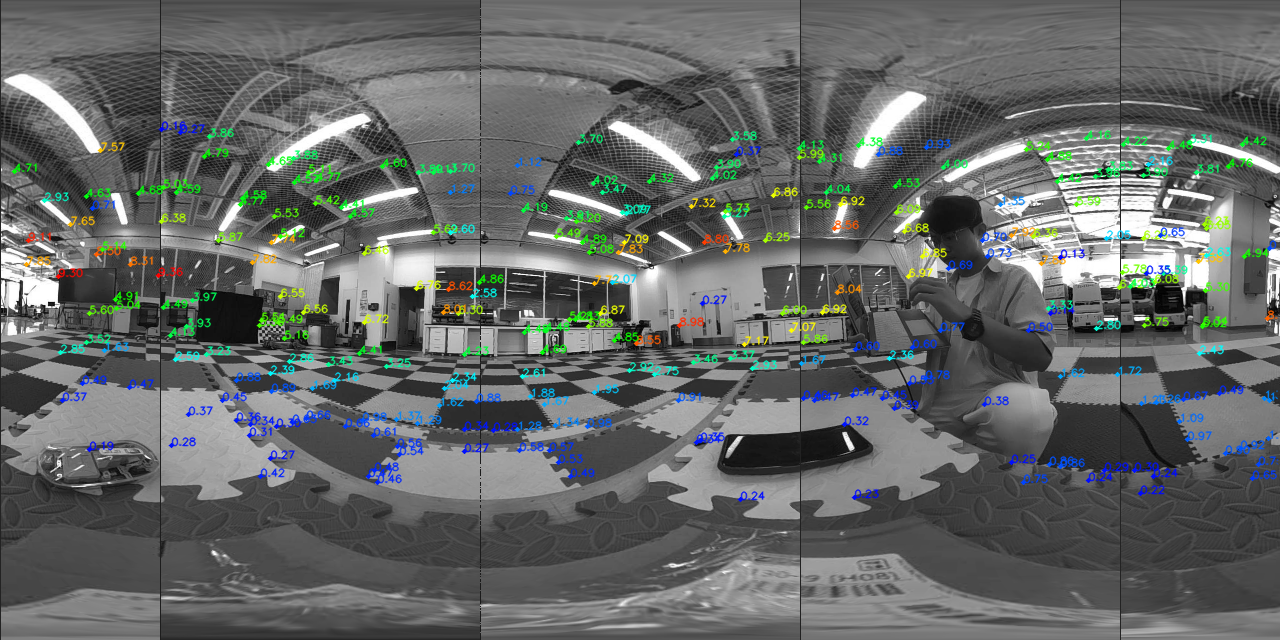}};
                      \node[
                        anchor=south west,       
                        font=\scriptsize,        
                        text=black,              
                        fill=yellow!40,          
                        inner sep=1.5pt,         
                        xshift=0.3em,            
                        yshift=0.3em             
                      ] at (img1.south west) {(a)};
                \end{tikzpicture}
                \hspace{-0.6em}
                \begin{tikzpicture}[inner sep=0pt, outer sep=0pt]
                    \node (img2) {\includegraphics[height=0.19\textheight]{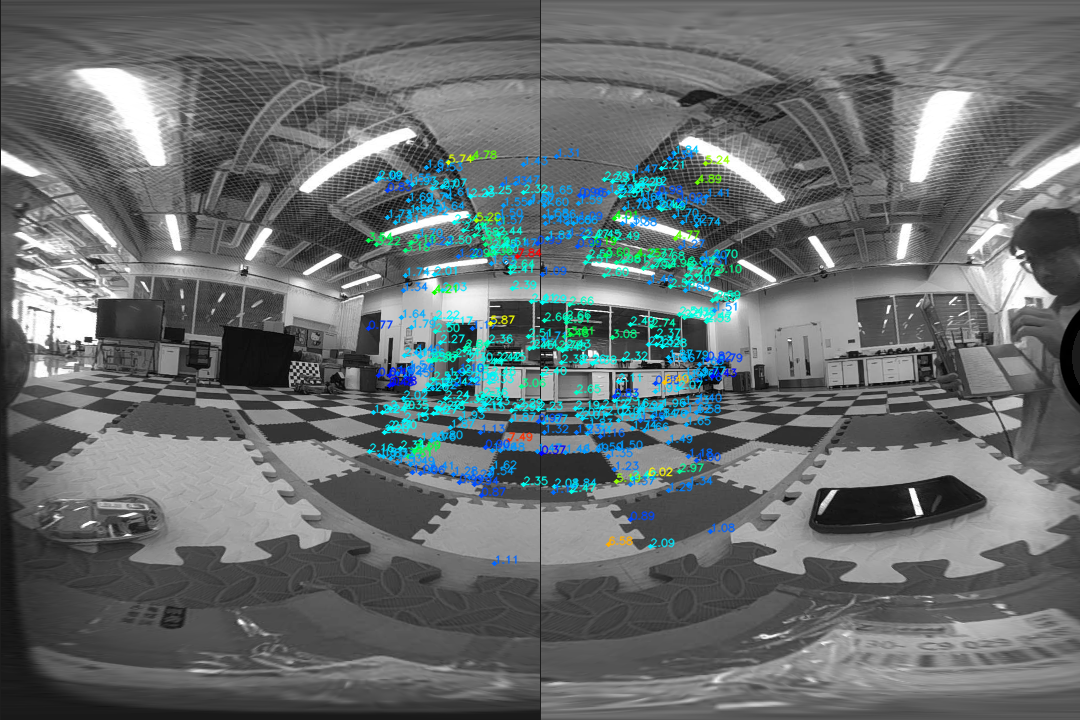}};
                      \node[
                        anchor=south west,       
                        font=\scriptsize,        
                        text=black,              
                        fill=yellow!40,          
                        inner sep=1.5pt,         
                        xshift=0.3em,            
                        yshift=0.3em             
                      ] at (img2.south west) {(b)};
                \end{tikzpicture}
                
                \vspace{0.3em} 
                
                \begin{tikzpicture}[inner sep=0pt, outer sep=0pt]
                    \node (img3) {\includegraphics[height=0.145\textheight]{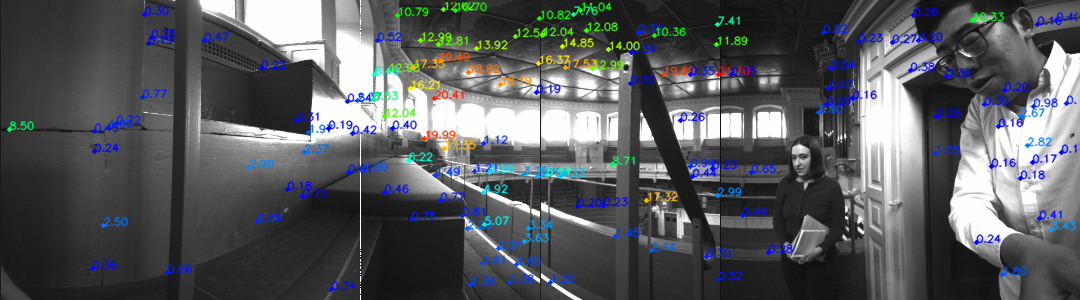}};
                     \node[
                        anchor=south west,       
                        font=\scriptsize,        
                        text=black,              
                        fill=yellow!40,          
                        inner sep=1.5pt,         
                        xshift=0.3em,            
                        yshift=0.3em             
                      ] at (img3.south west) {(c)};
                \end{tikzpicture}
                \hspace{-0.6em}
                \begin{tikzpicture}[inner sep=0pt, outer sep=0pt]
                    \node (img4) {\includegraphics[height=0.145\textheight]{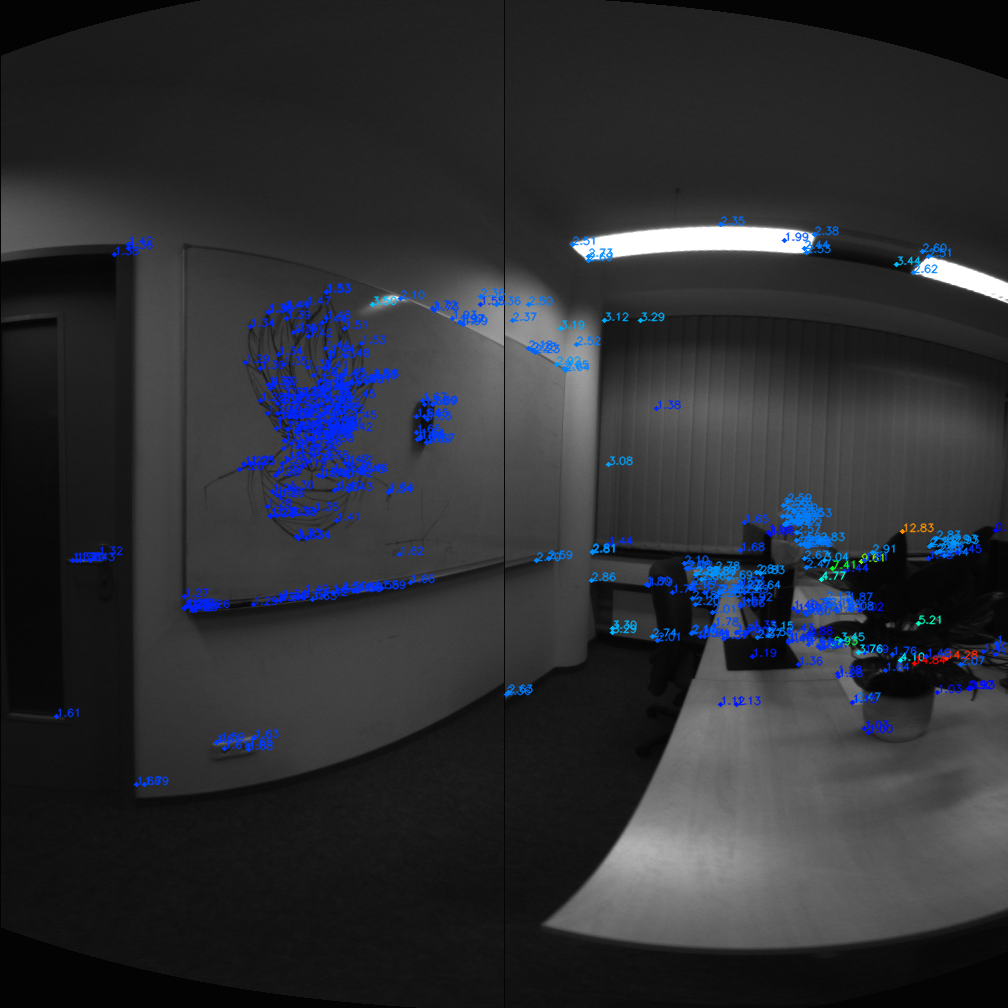}};
                     \node[
                        anchor=south west,       
                        font=\scriptsize,        
                        text=black,              
                        fill=yellow!40,          
                        inner sep=1.5pt,         
                        xshift=0.3em,            
                        yshift=0.3em             
                      ] at (img4.south west) {(d)};
                \end{tikzpicture}
            \end{minipage}
        };
        
        \node[
            rotate=90, 
            anchor=center, 
            font=\bfseries,
            xshift=0em, 
            yshift=\l_title_x
        ] at ({triangulation_block.center}) {Spherical Triangulation};
    \end{tikzpicture}
    
    \caption{Camera configurations and spherical triangulation results of four setups. The top row shows FOV coverage of four camera setups. Subfigures (a)–(d) correspond to Seeker Omnidirection, Seeker Stereo, HILTI 2022 and TUM-VI Stereo in sequence. Colored dots and text labels denote valid triangulated features, where colors encode estimated depth values (blue for near, red for far, unit: meters). The numerical labels next to each dot represent the corresponding estimated depth values.}
    \label{fig:spherical_triangulation}
\end{figure*}
\begin{table}[htbp]
\centering
\caption{ATE RMSE (m) on HILTI 2022 datasets \cite{hilti_dataset}. e*: Experiment*. Other evaluation details as in Table \ref{tab:euroc_result}.}
\label{tab:hilti2022_result}
\setlength{\tabcolsep}{0.5mm} 
\begin{tabular}{l|c|c|cccccc|c}
\toprule[1.0pt]
\multirow{1}{*}{Method}
& Bk
& \multirow{1}{*}{Cm}  
& e04 & e05 & e06 & e14 & e16 & e18 & \multirow{1}{*}{Avg} \\
\midrule[0.6pt] 
ORB-SLAM3\tnote{1}\,\,
& G & S 
& \num{0.232929} & \num{0.245078} & \num{0.223828} & \colorthird\num{0.200577} & \num{0.591487} & \num{0.31948} & \num{0.302229833} \\
MAVIS\tnote{1}\,\, 
& G & M 
& \colorfirst\num{0.072799} & \colorfirst\num{0.038883} & \colorfirst\num{0.085082} & \colorfirst\num{0.112977} & \colorfirst\num{0.105242} & \colorfirst\num{0.173012} & \colorfirst\num{0.097999167} \\
VINS-Fusion\tnote{1}\,\, 
& G & S 
& \num{0.342344} & \colorthird\num{0.194367} & \colorsecond\num{0.101541} & \num{0.308628} & \colorthird\num{0.467734} & \num{0.2733} & \colorthird\num{0.281319}  \\
\midrule[0.3pt] 
SchurVINS\tnote{1}\,\,
& F & S 
& \colorsecond\num{0.169894} & \num{0.220908} & \num{0.461713} & \num{0.286343} & \num{1.081387} & \colorsecond\num{0.246053} & \num{0.411049667} \\
SV (ours)\tnote{2}\,\,
& F & M 
& \colorthird\num{0.17249} & \colorsecond\num{0.124717} & \colorthird\num{0.169894} & \colorsecond\num{0.144499} & \colorsecond\num{0.231861} & \colorthird\num{0.248869} & \colorsecond\num{0.182055} \\
\bottomrule[1.0pt]
\end{tabular}
\end{table}

\subsection{Ablation Experiment} \label{sec:exp03_ablation}
\label{sec:ablation_experiment}
\subsubsection{Comparison on Omnidirectional Multi-Camera Dataset}
\label{sec:selfmade_omnidirectional_results}
To validate Sphere-VIO on real-world omnidirectional camera systems, we collected a custom multi-fisheye dataset with OptiTrack ground truth (Fig. \ref{fig:seeker_omni_d_photo}). As shown in Table \ref{tab:seeker_result}, all baselines fail: ORB-SLAM3 and MAVIS cannot adapt to over \ang[round-mode=none]{180} FOV due to their pinhole or Kannala-Brandt models, while VINS-Fusion and SchurVINS, despite using \ang[round-mode=none]{206} FOV-capable MEI models, fail due to cross-camera matching challenges from divergent orientations and partial overlaps.
\begin{table}[htbp]
\centering
\caption{ATE RMSE (m) on self-made datasets. s*: Sequences; NUC: Intel NUC 13 embedded. \textbf{Bold}: best average result. \textit{-} : No valid results available, due to tracking loss, or camera model incompatibility. Other details as in Table \ref{tab:euroc_result}.}
\label{tab:seeker_result}
\setlength{\tabcolsep}{0.7mm} 
\begin{tabular}{l|c|c|ccccc|c}
\toprule[1.0pt] 
\multirow{1}{*}{Method}
& Bk & Cm
& s01 & s02 & s03 & s04 & s05  & \multirow{1}{*}{Avg} \\
\midrule[0.6pt] 
VINS-Fusion\tnote{1}\,\, 
& G & S
& \textit{-} & \textit{-} & \textit{-} & \textit{-} & \textit{-} & \textit{-}  \\
ORB-SLAM3\tnote{1}\,\,
& G & S
& \textit{-} & \textit{-} & \textit{-} & \textit{-} & \textit{-} & \textit{-}  \\
MAVIS\tnote{1}\,\, 
& G & M
& \textit{-} & \textit{-} & \textit{-} & \textit{-} & \textit{-} & \textit{-}  \\
\midrule[0.3pt] 
SchurVINS\tnote{1}\,\,
& F & S
& \textit{-} & \textit{-} & \textit{-} & \textit{-} & \textit{-} & \textit{-}  \\
SV \tnote{2}\,\,
& F & S
& \num{0.161807} & \num{0.192095} & \num{0.123229} & \num{0.204328} & \num{0.216963} & \num{0.1796844}  \\
SV \tnote{2}\,\,
& F & M
& \num{0.123134} & \num{0.125983} & \num{0.146309} & \num{0.162568} & \num{0.05684} & \textbf{\num{0.1229668}}  \\
SV(NUC) \tnote{2,3}\,\,
& F & M
& \num{0.155513} & \num{0.139523} & \num{0.151597} & \num{0.146476} & \num{0.077356} & \num{0.134093}  \\
\bottomrule[1.0pt] 
\end{tabular}
\end{table}
In contrast, Sphere-VIO tracks stably on all sequences: the stereo variant achieves 0.1797 m average ATE RMSE, while the multi-camera version improves to 0.1230 m by leveraging overlapping coverage. On Intel NUC 13, performance degrades slightly to 0.1341 m, confirming resource-constrained deployment feasibility. SE(3)-aligned ATE RMSE is slightly higher than on public datasets due to two hardware limitations: a 5 cm stereo baseline, half that of standard datasets, degrading triangulation accuracy, and occasional duplicate IMU timestamps introducing noise. Despite these challenges, our method outperforms all baselines and aligns closely with ground truth (Fig. \ref{fig:trajectory_viewer}).

\subsubsection{Spherical Triangulation Across Diverse Camera Setups}
\label{sec:spherical_triangulation}
Fig. \ref{fig:spherical_triangulation} shows our initialization method's spherical triangulation results across four camera configurations with decreasing horizontal FOV: Seeker Omnidirection, HILTI 2022, Seeker Stereo, and TUM-VI Stereo. Direct comparison between Seeker Omnidirection and Stereo reveals the root cause of the performance gap in previous system-level ablations. The stereo configuration achieves valid static triangulation only in the central \ang[round-mode=none]{90} horizontal FOV, covering just one-quarter of the full omnidirectional panorama, while features outside this region cannot be statically initialized and rely on depth estimation during VIO tracking, introducing uncertainty and degrading accuracy. In contrast, the four-camera configuration achieves uniform 360° full-coverage triangulation, providing a more robust initialization foundation. Consistent results on HILTI 2022 and TUM-VI Stereo further demonstrate our method's generalization to diverse multi-camera layouts and the universality of the USPM framework.
\subsection{Computational Efficiency On Multi-Camera Dataset} \label{sec:exp04_efficiency}
\label{sec:processing_time_experiment}
We evaluate per-frame latency on the HILTI 2022 dataset. For fairness, VINS-Fusion is configured to run backend optimization on every frame, matching the full-frame front-end and back-end pipeline of filter-based methods (SchurVINS and Sphere-VIO). MAVIS and ORB-SLAM3 are excluded due to their complex multi-threaded architectures preventing accurate latency measurement, leaving VINS-Fusion as the graph-based baseline. With 40 Hz cameras, HILTI 2022 requires per-frame latency $\leq$ 0.025 s for real-time operation. As shown in Table \ref{tab:hilti2022_realtime_performance}, stereo VINS-Fusion (0.0257 s) exceeds this limit, while SchurVINS and Sphere-VIO run in real time. Our multi-camera Sphere-VIO, processing four streams with CLAHE preprocessing, averages only 0.0122 s per frame, nearly half the threshold. This efficiency stems from our three parallel units implemented via a thread pool (Section \ref{subsec:universal_camera_model}), fully utilizing multi-core CPUs. Then as listed in Table \ref{tab:seeker_realtime_performance}, on our 20 Hz self-made dataset (latency limit 0.05 s, Section \ref{sec:selfmade_omnidirectional_results}), Sphere-VIO achieves 0.0138 s on a laptop and 0.0180 s on Intel NUC 13. Both are well within the limit, verifying practical deployment feasibility.
\begin{table}[htbp]
\centering
\caption{Per-frame processing time (in seconds) on the HILTI 2022 dataset \cite{hilti_dataset}. Other details as in Table \ref{tab:seeker_result}.}
\label{tab:hilti2022_realtime_performance}
\setlength{\tabcolsep}{0.5mm} 
\begin{tabular}{l|c|c|cccccc|c}
\toprule[1.0pt] 
Method
& Bk & Cm
& e04 & e05 & e06 & e14 & e16 & e18 & \multirow{1}{*}{Avg} \\
\midrule[0.6pt] 
\multirow{1}{*}{VINS-Fusion\tnote{1}\,\,} 
& G & S
& \num{0.02775} & \num{0.025435} & \num{0.024518} & \num{0.02322} & \num{0.026321} & \num{0.027165} & \num{0.025734833} \\
\midrule[0.3pt] 
\multirow{1}{*}{SchurVINS\tnote{1}\,\,}
& F & S 
& \num{0.003659} & \num{0.004364} & \num{0.00385} & \num{0.003243} & \num{0.003102} & \num{0.003382} & \textbf{\num{0.0036}} \\
\multirow{1}{*}{SV (ours)\tnote{2}\,\,}
& F & M 
& \num{0.013355} & \num{0.012959} & \num{0.012574} & \num{0.011106} & \num{0.01147} & \num{0.011597} & \num{0.012176833} \\
\bottomrule[1.0pt] 
\end{tabular}
\end{table}
\begin{table}[htbp]
\centering
\caption{Per-frame processing time (in seconds) on the self-made dataset. Other details as in Table \ref{tab:seeker_result}.}
\label{tab:seeker_realtime_performance}
\setlength{\tabcolsep}{1mm} 
\begin{tabular}{p{1.5cm}|c|c|ccccc|c}  
\toprule[1.0pt] 
Method
& Bk & Cm
& s01 & s02 & s03 & s04 & s05  & Avg \\
\midrule[0.6pt] 
SV\tnote{1}\,  
& F & M
& \num{0.013534} & \num{0.013732} & \num{0.013458} & \num{0.014798} & \num{0.013597} & \textbf{\num{0.0138238}}  \\
SV(NUC)\tnote{1,2}\,  
& F & M
& \num{0.016455} & \num{0.017794} & \num{0.018056} & \num{0.019519} & \num{0.017942} & \num{0.0179532}  \\
\bottomrule[1.0pt] 
\end{tabular}
\end{table}
\section{CONCLUSION AND FUTURE WORK}
This paper presents Sphere-VIO, a CPU-only filter-based panoramic VIO framework for multi-camera wide-FOV fisheye systems. With its novel panoramic global management paradigm and spherical residual-based ESKF, it adapts seamlessly to diverse camera setups and enables robust state estimation. Extensive evaluations show Sphere-VIO achieves competitive accuracy, superior robustness, and reliable real-time performance on embedded hardware without GPU dependency. It effectively balances generality, efficiency, and deployability for resource-constrained robots. As a pure VIO system, it lacks loop closure and suffers from long-term drift. Future work will incorporate lightweight loop closure to extend it to a complete SLAM system while preserving its CPU-only efficiency and multi-camera adaptability.

\addtolength{\textheight}{-12cm}   









\bibliographystyle{IEEEtran}  
\bibliography{IEEEabrv,sphere_vio_ref}         

\end{document}